\documentclass{article}

\usepackage{microtype}
\usepackage{graphicx}
\usepackage{booktabs} % for professional tables
\usepackage{xspace}
\usepackage{tikz}
\usepackage{mwe}
\usepackage{amsfonts}
\usepackage{subcaption}
\usepackage{amsmath}
\usepackage{multirow}
\usepackage{array}
\usepackage{enumitem}
\newcommand{\tinyskip}{\vspace{3pt}}
\newcommand{\mypar}[1]{\tinyskip\noindent\textbf{#1.}\xspace}

\usepackage{hyperref}

\newcommand{\appref}[1]{Appendix~\ref{#1}}

\usepackage[accepted]{mlsys2025}

\mlsystitlerunning{ContextPilot: Fast Long-Context Inference via Context Reuse}
\begin{document}
\begin{sloppypar}
\twocolumn[
\mlsystitle{ContextPilot: Fast Long-Context Inference via Context Reuse}

% unless you've provided the [accepted] option to the mlsys2025


\mlsyssetsymbol{equal}{*}
\begin{mlsysauthorlist}
\mlsysauthor{Yinsicheng Jiang}{equal,uoe}
\mlsysauthor{Yeqi Huang}{equal,uoe}
\mlsysauthor{Liang Cheng}{uoe}
\mlsysauthor{Cheng Deng}{uoe}
\mlsysauthor{Xuan Sun}{uoe}
\mlsysauthor{Luo Mai}{uoe}
\end{mlsysauthorlist}

\mlsysaffiliation{uoe}{University of Edinburgh}
% \mlsysaffiliation{tencent}{Tencent}
% \mlsysaffiliation{ed}{School of Computation, University of Edenborrow, Edenborrow, United Kingdom}

\mlsyscorrespondingauthor{Yinsicheng Jiang}{ysc.jiang@ed.ac.uk}
\mlsyscorrespondingauthor{Yeqi Huang}{yeqi.huang@ed.ac.uk}
\mlsyscorrespondingauthor{Liang Cheng}{L.cheng@ed.ac.uk}
\mlsyscorrespondingauthor{Cheng Deng}{cdeng@ed.ac.uk}
\mlsyscorrespondingauthor{Xuan Sun}{xuan.sun@ed.ac.uk}
\mlsyscorrespondingauthor{Luo Mai}{luo.mai@ed.ac.uk}

\mlsyskeywords{Long-Context LLM Inference, Retrieval-Augmented Generation (RAG), Agentic AI}

\vskip 0.3in

\begin{abstract}
AI applications increasingly depend on long-context inference, where LLMs consume substantial context to support stronger reasoning. Common examples include retrieval-augmented generation, agent memory layers, and multi-agent orchestration. As input contexts get longer, prefill latency becomes the main bottleneck. Yet today’s prefill acceleration techniques face a trade-off: they either preserve reasoning quality but deliver little KV-cache reuse, or improve reuse at the cost of degraded reasoning quality. 

We present \textsc{ContextPilot}, a system that accelerates prefill by introducing \emph{context reuse} as a new mechanism for faster long-context inference. \textsc{ContextPilot} introduces a context index to identify overlapping \emph{context blocks} across LLM interactions (e.g., across users and turns). It further proposes context alignment and de-duplication techniques to maximize KV-cache reuse. To preserve reasoning quality under reuse, it introduces succinct context annotations that prevent quality degradation. Finally, \textsc{ContextPilot} is built around a modular architecture with a clean interface that integrates with existing inference engines. Extensive evaluation shows that \textsc{ContextPilot} reduces LLM prefill latency by up to 3\texttimes{} compared to state-of-the-art methods while preserving reasoning quality. At longer context lengths, it can even improve reasoning quality. \textsc{ContextPilot} is open-sourced at: \url{https://github.com/EfficientContext/ContextPilot}.
\end{abstract}
]



\printAffiliationsAndNotice{\mlsysEqualContribution} % otherwise use the standard text.
\section{Introduction}

Long-context inference is now central to many AI applications. Whether through retrieval-augmented generation (RAG)~\cite{lewis2020rag}, AI memory layers such as Mem0~\cite{mem0}, multi-agent orchestration, or personal AI assistants that interact with external data across conversations (e.g., OpenClaw), modern workloads routinely feed LLMs tens to hundreds of thousands of tokens of external context. In a typical pipeline, a retriever (e.g., FAISS, Qdrant, ElasticSearch), memory store, or agent tool call (e.g., file reading, web search) fetches relevant documents, chunks, or memories for a user query, and an inference engine (e.g., SGLang, vLLM, TensorRT-LLM) consumes them as input context.

We call these discrete units of external context \emph{context blocks (CBs)}. During the \emph{prefill} phase, the engine computes key–value (KV) caches, which are then reused during \emph{decode} to generate output tokens sequentially. The key performance goal in prefill is to reduce time-to-first-token (TTFT). To that end, inference engines use a \emph{prefix cache} that stores KV caches from prior prompts, avoiding recomputation for repeated inputs or prompts that share a prefix.

AI applications are feeding LLMs ever larger amounts of external context to unlock stronger capabilities, making prefill latency a key bottleneck. This trend is driven by two forces. First, many studies show that more context—ranging from retrieved knowledge-base content to long-horizon memories in agentic systems—improves performance on complex reasoning tasks (e.g., lemmas in AI4Math)~\cite{hilbert}, strengthens access to up-to-date information (e.g., AI4Search)~\cite{opendeepsearch, deepsearch, deepsearchqa}, and reduces hallucinations~\cite{Ayala_2024, RAReducehallu, 11014810}. Second, while chunking is widely used to shrink per-request inputs, recent findings~\cite{anthropic2025contextengineering} suggest that overly aggressive chunking can harm reasoning quality; in contrast, processing larger, less fragmented context (e.g., full documents) often yields better results—further increasing prefill latency.

To speed up long-context inference, existing systems adopt two main techniques, yet each faces a trade-off between reuse efficiency and model accuracy.
The first, \emph{exact prefix matching}, used in systems such as RadixCache~\cite{sglang}, LMCache~\cite{lmcache}, and RAGCache~\cite{ragcache}, reuses cached KV states only when a new prompt exactly matches a previous prefix.
This approach preserves accuracy but yields low cache-hit ratios in practice, as long-context workloads often retrieve large sets of documents or memories in varying orders, leaving most KV caches unused.
The second category, \emph{approximate KV-cache matching}, exemplified by CacheBlend~\cite{cacheblend} and PromptCache~\cite{prompcache}, matches KV caches by floating-point similarity rather than exact prefixes.
While this increases reuse and shortens TTFT, we observe in evaluation that it can significantly degrade model accuracy.

To reduce TTFT for long-context inputs without sacrificing accuracy, we propose a new approach based on the observation that real-world long-context workloads often exhibit overlapping context blocks, commonly (i) across multiple turns within the same conversation and (ii) among parallel sessions (e.g., prompts or user queries) in domain-specific applications. 
Leveraging this observation, we identify three opportunities for \emph{context reuse with negligible accuracy loss}: 
(1) \emph{Aligning} context blocks with previously cached prefixes, improving cache-hit ratios;
(2) \emph{De-duplicating} context blocks to avoid recomputation for already cached content; and 
(3) \emph{Adding context annotations} to inform the model of original relevance ranking and deduplicated block locations, mitigating accuracy loss.

In this paper, we present \textsc{ContextPilot}, a system that accelerates prefill by introducing \emph{context reuse} as a new mechanism for faster long-context inference. \textsc{ContextPilot} targets practical long-context settings with parallel sessions and multi-turn conversations, where substantial portions of the input context recur across requests. Our key contributions are summarized below.

\mypar{(1) Context Indexing}
We design an indexing mechanism that efficiently tracks cached context blocks across parallel sessions and multi-turn conversation histories.
The index supports fast retrieval of previously stored contexts by (i) aligning prefix overlaps between the incoming context and cached blocks, and (ii) traversing blocks referenced in multi-turn histories to recover reusable segments beyond the immediate prefix match, while maintaining low construction and maintenance overhead.

\mypar{(2) Context Alignment}
We propose a context alignment algorithm that queries the index to align context blocks with the prefix cache, with the explicit goal of maximizing cache hit ratio under a fixed context budget.
To mitigate potential accuracy loss introduced by aligning, we introduce concise \emph{order annotations} that preserve the original relevance ranking and structural cues, allowing the LLM to interpret the aligned prompt consistently with the intended semantics.

\mypar{(3) Context De-Duplication}
We further improve reuse efficiency via context de-duplication.
By querying the index, \textsc{ContextPilot} identifies context blocks that overlap with already cached contexts (including partial overlaps), and replaces duplicated spans with succinct \emph{location annotations} that point to their original occurrences in the prompt (or prior turns), thereby avoiding redundant prefill while preserving accuracy.

Extensive evaluations show that \textsc{ContextPilot} delivers strong performance across diverse baselines and real-world datasets. Across long-context workloads---including RAG (multi-turn, multi-session, and hybrid), agentic memory systems (Mem0), emerging multi-agent reasoning paradigms, and real-world agent deployments (OpenClaw)---it reuses contexts to accelerate prefill, outperforming state-of-the-art systems (CacheBlend, LMCache, RadixCache, and RAGCache) by 1.5--3$\times$ on MultihopRAG, NarrativeQA, QASPER, and MT-RAG with negligible accuracy loss. As context length grows, \textsc{ContextPilot} can even improve reasoning quality and answer accuracy, thanks to its novel context-annotation design.

\textsc{ContextPilot} also scales to very large MoE models~\cite{moecap}: on DeepSeek-R1 (671B), it improves prefill throughput by 1.52--1.81$\times$ on 16--32 GPUs.
Beyond cloud deployments, \textsc{ContextPilot} reduces prefill latency by 63.6\% in a real-world agent pipeline (OpenClaw on a single RTX~5090), and achieves ${\sim}2.4\times$ latency reduction on Apple Silicon laptops, demonstrating broad applicability from data center to edge.

Building on these results, we are working towards broader academic and industry deployment with multiple adopters, and have open-sourced \textsc{ContextPilot} on GitHub. We expect it to be deployed in more challenging multi-user serverless scenarios~\cite{Fu2024ServerlessLLM} and can serve as an extensible software foundation for context engineering~\cite{hua2025context,contextengineeringsurvey2025}, replay~\cite{liu2025contextual}, management~\cite{chang2025sagallm,memagent}, and optimization~\cite{kang2025acon,li2025dbperspective}.

\section{Background and Motivation}

\subsection{Long-context inference systems}

Long-context inference systems augment LLMs with external context blocks\textemdash retrieved documents, chunks, or memories\textemdash to enhance factual grounding and reasoning.
We use the term \emph{context block} throughout this paper to refer to any discrete unit of external context injected into the model, whether a retrieved document, a document chunk, or a memory entry.
Two dominant paradigms drive this trend:
(1) \emph{Retrieval-augmented generation (RAG)}~\cite{lewis2020rag,gao2024ragsurvey} retrieves the top-$K$ most relevant documents per query from an external corpus, serving both online latency-sensitive services (e.g., semantic search, dialogue, deep research~\cite{deepsearch, guo2024ds}) and offline throughput-oriented pipelines (e.g., large-scale annotation, synthetic data generation~\cite{ragsynth, megapairs, nvidia-post, culturesynth}).
(2) \emph{AI memory layer} systems (e.g., Mem0~\cite{mem0}) dynamically extract, consolidate, and retrieve user-specific memories across sessions~\cite{memorysurvey2025}, injecting relevant context blocks into each query to enable personalized interactions.
In both cases, the inference engine performs \emph{prefill} to encode these context blocks and \emph{decode} to generate responses.

\begin{figure}[!t]
    \centering
    \includegraphics[width=\linewidth]{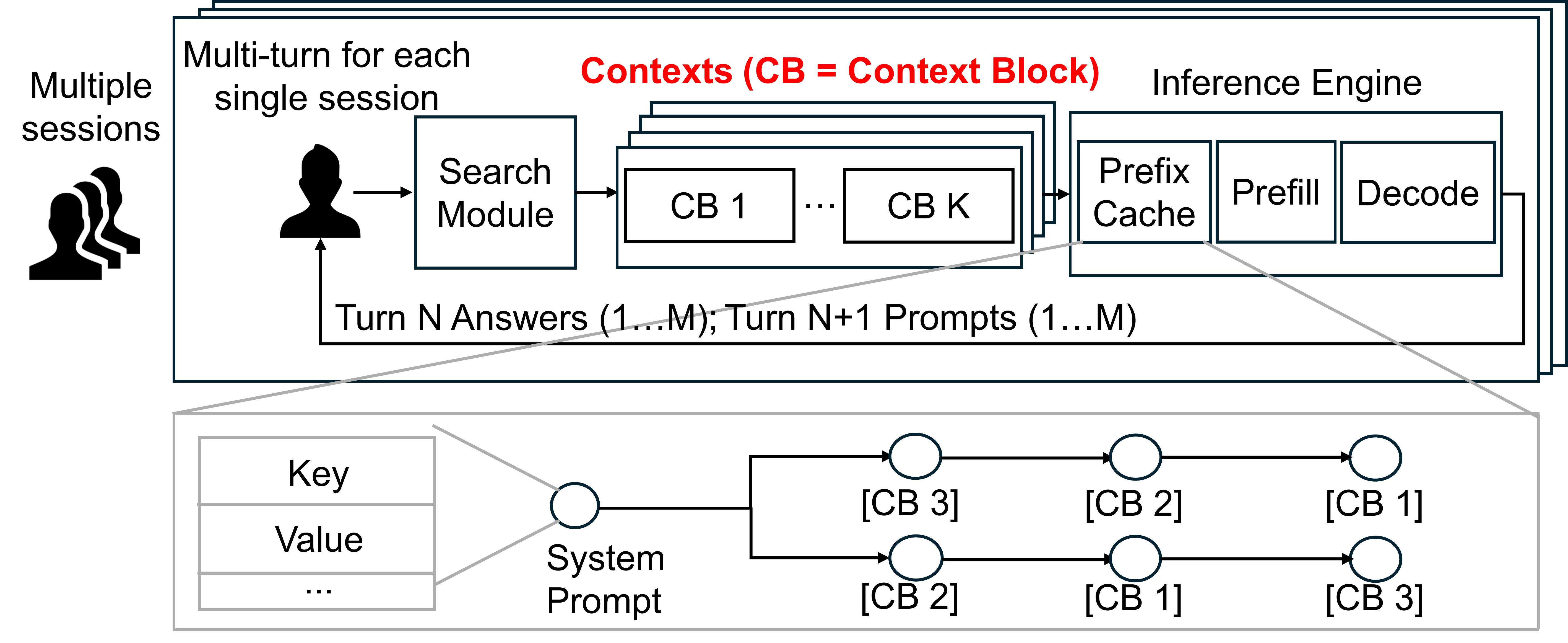}
    \vspace{-0.8cm}
    \caption{Overview of a long-context inference system with prefix caching.}
    \vspace{-0.8cm}
    \label{fig:overview}
\end{figure}

A typical system (~\autoref{fig:overview}) alternates between retrieval (or memory lookup) and generation across sessions and dialogue turns. 
At each turn, $M$ concurrent prompts receive $K$ relevant context blocks each. 
The inference engine encodes the context and generates responses, which feed into the next step with updated dialogue history, enabling efficient multi-turn reasoning.

These long-context inference systems often use prefix caching to improve prefill efficiency~\cite{dontbreakcache}.
A trie-based implementation~\cite{sglang} organizes tokens hierarchically, with each node storing a token sequence and its KV cache, enabling longest-prefix matching through a single traversal.
An alternative hash-table design~\cite{vllm} directly maps complete prefixes to KV-block identifiers.

% Yet hash computations scale with input length and concurrency, and can introduce high overhead under low cache-hit ratios.

% Additionally, block-level granularity prevents partial caching, causing redundancy when concurrent requests have overlapping but non-identical prefixes.

% RAG mitigates two fundamental issues: temporal staleness (e.g., evolving source code or new publications) and data isolation (e.g., restricted internal documents or user-specific records). By retrieving semantically relevant information from external corpora, RAG dynamically extends the effective context window of an LLM~\cite{shi2023replug}, enabling grounded and verifiable responses beyond its pretraining distribution. 

\subsection{Emerging challenge: growing context lengths}

Long-context inference systems face a critical prefill latency bottleneck as modern LLMs demand expanding context windows. This is driven by two reasons: (1)~\emph{increasing the number of retrieved context blocks} to broaden information coverage~\cite{li2024retrieval,jin2024longcontextllmsmeetrag,yueinference,laban2024summary,chung2025longcontext}, and~(2) enriching contextual information by \emph{retrieving complete documents or full memory histories} and applying context engineering methods~\cite{anthropic2025contextengineering}.

Analysis of our workload data reveals that both approaches deliver significant accuracy gains. Scaling the retrieval parameter ($k$) from lower to higher values enhances accuracy by as much as 20\%, while retrieving full documents achieves similar performance improvements, confirmed by recent context engineering studies~\cite{zhang2025agenticcontextengineeringevolving}.

However, expanded context windows (i.e., longer context block inputs) introduce substantial prefill overhead and can even degrade reasoning quality beyond a certain length~\cite{du2025contextlength,raju2026limits}. Our trace data shows that LLM inference engines often process 20k--130k prefill tokens, leading to 3--10 second latency when executing 32B dense models on a single H100 GPU. For larger models such as Mixture-of-Experts (MoEs), the prefill latency can be even higher.
As a result, the prefill becomes the dominant bottleneck, downgrading user experience and preventing long-context applications from being widely deployed.

\subsection{Issues of existing KV cache reuse methods}

To address the growing cost of longer retrieved contexts, existing KV-cache reuse methods exhibit several issues:

\mypar{Exact-prefix matching yields low KV-cache reuse} 
Existing prefix-caching mechanisms rely heavily on exact token-level matching, e.g., RadixCache~\cite{sglang}, or document-level matching, e.g., LMCache~\cite{lmcache} and RAGCache~\cite{ragcache}: even minor variations, such as whitespace differences or slightly reordered tokens and documents, prevent reuse. Our evaluation (\autoref{subsec:eval-app}) shows that despite substantial overlap in retrieved documents across related queries, cache hit ratios remain persistently low. For example, for the dataset multihopRAG with Qwen3-32B, the KV-cache hit ratio is only 4.6\%, indicating low KV cache reuse. For NarrativeQA with Llama3.3-70B, the hit ratio is also only 5.5\%, leaving most cache unused.

\mypar{Approximate KV-cache matching degrades quality}
To improve low cache-hit ratios, recent techniques such as CacheBlend~\cite{cacheblend} adopt approximate KV-cache matching. Instead of exact-prefix matching, they measure similarity in KV values (floating-point vectors) and reuse cached states when the proximity exceeds an empirically decided threshold. However, KV-value similarity is \emph{not} a reliable indicator of whether cached states can be reused across different contexts and requests. Approximate matching degrades accuracy, with errors compounding over multi-turn interaction. Our evaluations (\autoref{subsec:eval-app}) show that across multiple models (e.g., Qwen3-32B, Qwen3-4B, Llama3.3-70B) and datasets (e.g., MultihopRAG, NarrativeQA, QASPER), approximate matching can degrade accuracy by 9--11\% (dropping from around 60\% to approximately 50\%), preventing its deployment in many services where high fidelity is necessary.

% Nevertheless, this flexibility introduces fundamental accuracy-efficiency trade-offs that become pronounced in RAG workloads. The reasoning dependencies among retrieved documents exhibit complex, dynamic patterns that vary with query semantics and document content. Fixed recomputation policies fail to adequately capture these non-local semantic interactions, potentially compromising reasoning quality when cached approximations diverge from exact computations.

% Bridging retrieval semantics with cache mechanics remains an open challenge. Effective solutions require retrieval–cache co-design: retrieval systems must produce deterministic, stable document orderings, while caching mechanisms must support quality-aware matching. Realizing high cache efficiency without compromising reasoning accuracy demands principled integration of retrieval strategies with cache management policies.
\section{Design Overview}\label{sec:motivation}

\begin{figure}[!t]
    \centering
    \begin{subfigure}{0.48\linewidth}
        \centering
        \includegraphics[width=\linewidth]{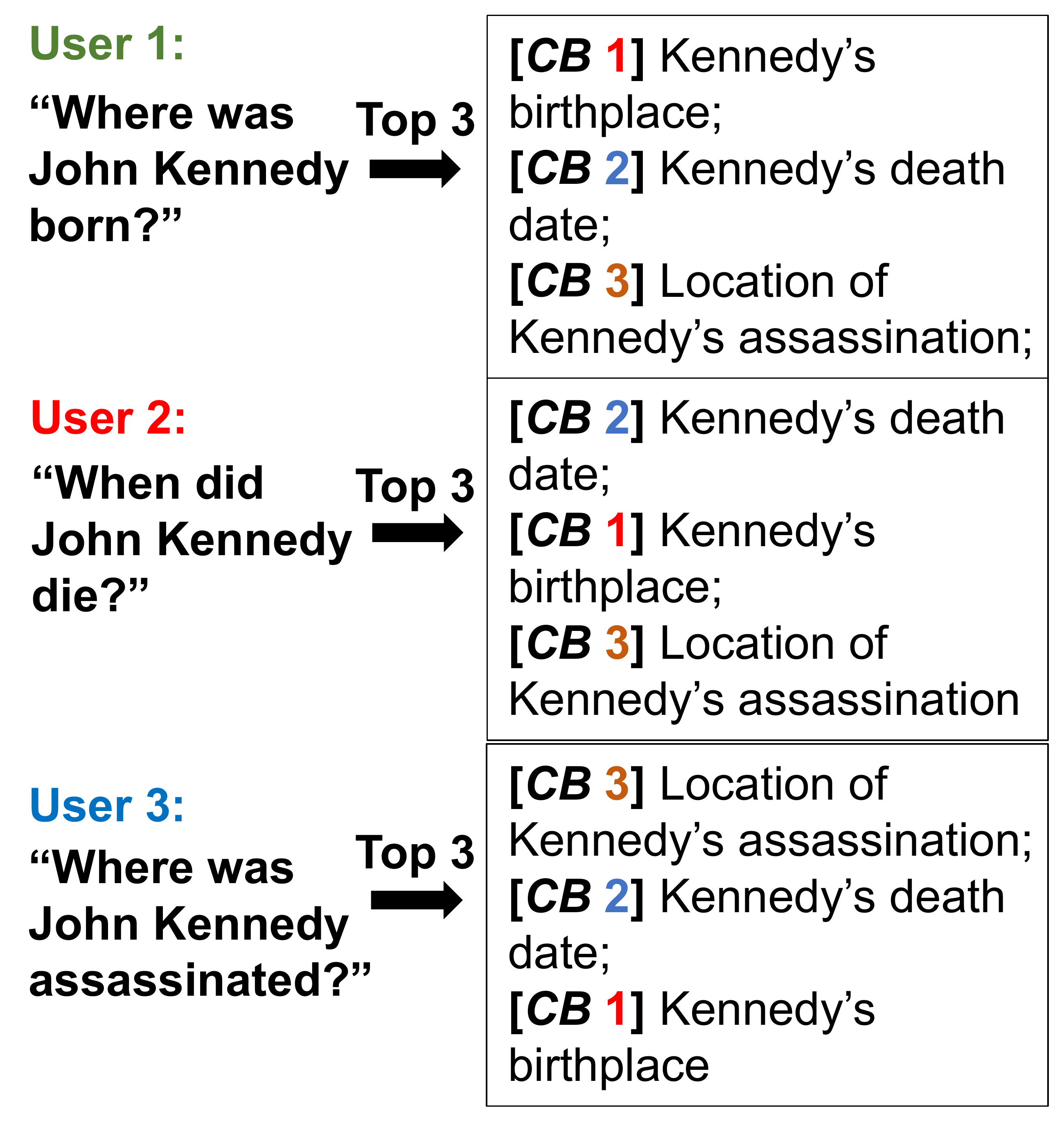}
        \caption{Multi-session overlap.}
        \label{fig:example_multi_session}
    \end{subfigure}
    \hfill
    \begin{subfigure}{0.48\linewidth}
        \centering
        \includegraphics[width=\linewidth]{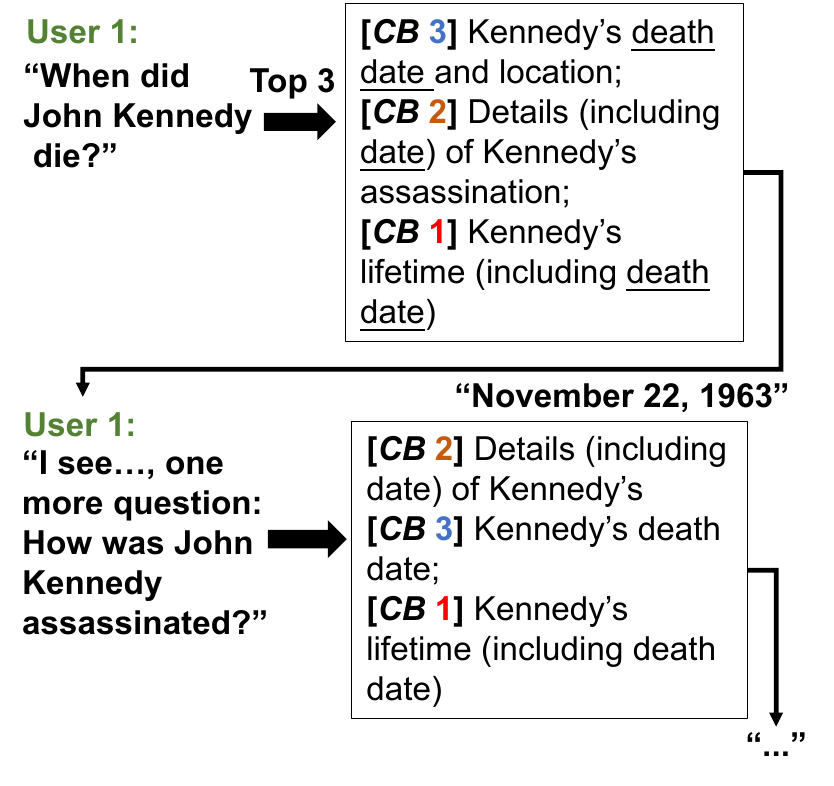}
        \caption{Multi-turn overlap.}
        \label{fig:example_multi-turn}
    \end{subfigure}
    
    \vspace{0.5em}
    
    \begin{subfigure}{1.\linewidth}
        \centering
        \includegraphics[width=\linewidth]{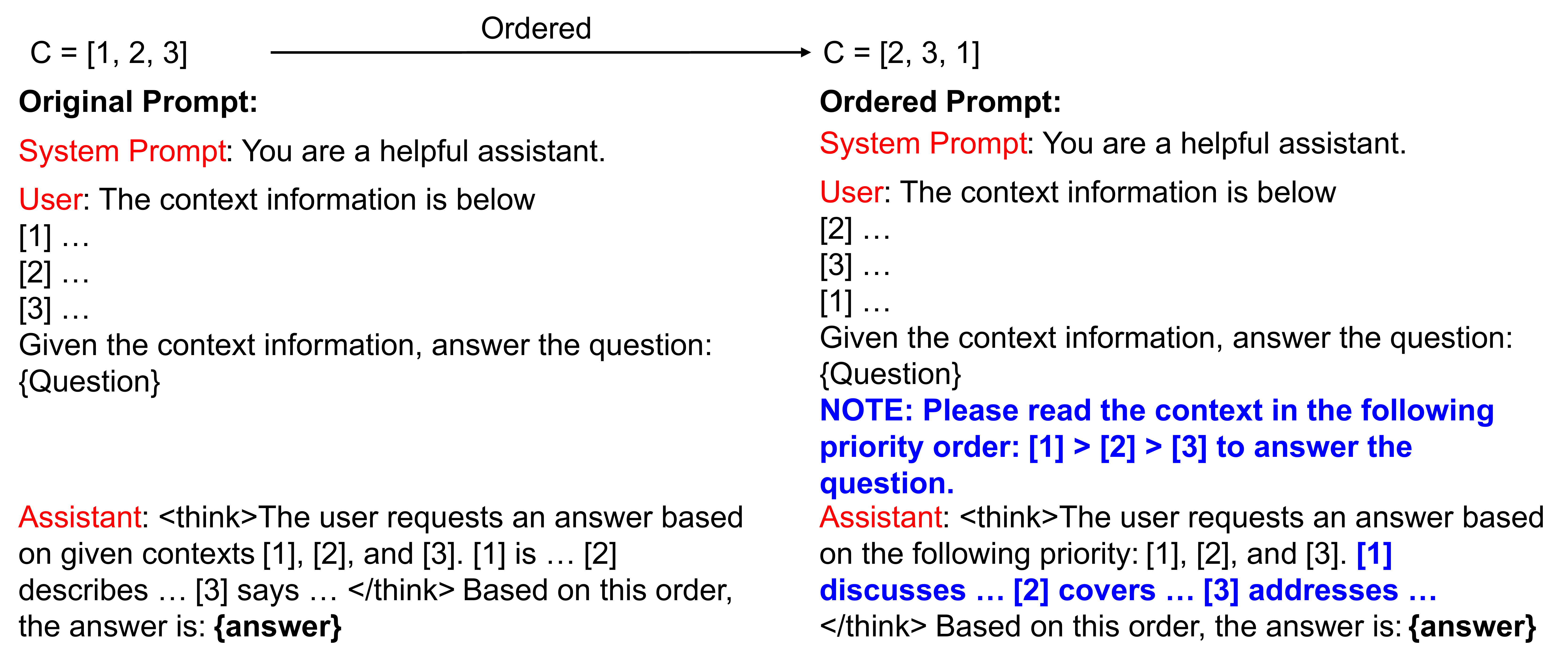}
        \caption{Context annotations for recovering relevance ranking semantics.}
        \label{fig:order_recovery}
    \end{subfigure}
    \caption{Context overlap and reuse opportunities in long-context inference.}
    \label{fig:Examples_of_overlaps}
    \vspace{-0.4cm}
\end{figure}

% \begin{table}[t]
%   \centering
%   \resizebox{\linewidth}{!}{%
%   \begin{tabular}{rccc}
%     \toprule
%     & \multicolumn{3}{c}{\textbf{Dataset}} \\
%     \cmidrule(lr){2-4}
%     \textbf{Top \% Documents} & \textbf{MultihopRAG} & \textbf{NarrativeQA} & \textbf{QASPER} \\
%     \textbf{(by frequency)} & \multicolumn{3}{c}{\textbf{(\% of Retrieval Events with $\geq$1 Hit)}} \\
%     \midrule
%     % 1  & 14.7\% & 11.5\% & 5.7\% \\
%     % 5  & 47.3\% & 25.8\% & 19.5\% \\
%     10 & 64.1\% & 37.7\% & 31.7\% \\
%     20 & 79.2\% & 57.4\% & 49.6\% \\
%     % 50 & 95.3\% & 88.7\% & 80.9\% \\
%     \bottomrule
%   \end{tabular}}
%   \vspace{-15pt}
%   % \parbox{0.95\linewidth}{\footnotesize
%   % A retrieval event is a single query returning $k$ documents. Documents are ranked globally by retrieval frequency within each dataset. Each entry shows the percentage of retrieval events that include at least one document from the corresponding top-$p\%$ most frequently retrieved documents.}
%     \caption{Retrieval concentration across datasets. Each entry shows the percentage of retrieval events that return at least one document from the top-$p\%$ most frequently retrieved documents in that dataset.}
%     \label{tab:overlap-retrievals}
% \end{table}

\subsection{Observation: significant overlap in long context}

Our design is motivated by a key observation: real-world long-context workloads exhibit substantial overlap in context blocks across both sessions and conversation turns:

\mypar{(1) Overlap across sessions}  
~\autoref{fig:example_multi_session} illustrates overlapping retrievals among multiple users querying different aspects of the same person. 
Although the retrieved documents appear in different orders reflecting per-query relevance, their content largely coincides. 
Trace studies on MultihopRAG~\cite{mulhoprag}, NarrativeQA~\cite{narrativeqa}, and QASPER~\cite{qasper} confirm this trend: 79.2\%, 57.4\%, and 49.6\% of questions respectively draw from the top 20\% most frequently accessed documents (\autoref{fig:cdf} in \appref{apd:cdf}), indicating extensive context sharing across sessions.

\mypar{(2) Overlap within multi-turn conversations}  
\autoref{fig:example_multi-turn} shows that in multi-turn interactions, users often revisit related topics, causing the retriever to return the same documents with slightly different rankings. 
As previous turns become part of the input, later retrievals frequently duplicate content already present in the cached history.
Our MT-RAG~\cite{mtrag} trace study quantifies this effect: on average, 40\% of retrieved documents in any turn overlap with earlier ones in the same session.
Even within a single turn, distinct context blocks can share overlapping content (e.g., Kennedy's death date appears across multiple blocks), creating content-level redundancy.

% \subsection{Design opportunities for context reuse}\label{subsec:design-opp}

\subsection{Design opportunities for context reuse}\label{subsec:design-opp}

The significant overlap among context blocks reveals clear opportunities for \emph{context reuse}, boosting KV-cache hit ratio. Specifically, we identify three opportunities that commonly arise in real-world long-context applications:

\mypar{(1) Aligning context blocks with the prefix cache across sessions boosts KV-cache reuse}
As shown in \autoref{fig:example_multi_session}, if the context blocks for the second and third users are aligned to match the first user's sequence, all three contexts would share an identical prefix, achieving 100\% KV-cache reuse.

Trace-based alignment experiments on MultihopRAG, NarrativeQA, and QASPER confirm this potential. 
Aligning context block order with prefix-cache structure raises KV-cache hit ratio to 38.9\%, 20.2\%, and 16.5\%, respectively, representing 3–8$\times$ higher utilization than the baseline (\autoref{subsec:eval-breakdown}).
Thus, strategic context alignment can dramatically cut redundant prefill computation across users.

Crucially, such alignment incurs minor accuracy loss: only 0.1–3.3\% on the same datasets (\autoref{subsec:eval-breakdown}).
As shown in \autoref{tab:demo-reproduce}, our reproduction of the DEmO ordering study~\cite{demo} with newer models confirms that modern LLMs are substantially less sensitive to input ordering than earlier generations, with near-zero variance on datasets (SST2~\cite{sst2}, SNLI~\cite{snli}, SUBJ~\cite{subj}, CR~\cite{cr}) that showed large gaps in the original study.
The small residual degradation arises because prefix-optimized alignment can occasionally move important context blocks toward the middle of the list, exposing them to the lost-in-the-middle effect~\cite{liu2023lostmiddlelanguagemodels}.

\begin{table}[t]
\centering
\caption{Reproducing DEmO ordering study with newer models. Modern LLMs show negligible ordering gaps even on datasets that showed large gaps in the original study.}
\label{tab:demo-reproduce}
\resizebox{\linewidth}{!}{%
\begin{tabular}{l|cc|cc}
\toprule
\multirow{2}{*}{\textbf{Dataset}} & \multicolumn{2}{c|}{\textbf{GPT-3.5}} & \multicolumn{2}{c}{\textbf{GPT-5.1}} \\
& Random & DEmO & Random & DEmO \\
\midrule
SST2~\cite{sst2} & 93.8 & 93.8 & 92.0 & 93.8 \\
SNLI~\cite{snli} & 72.6 & 72.6 & 83.2 & 83.2 \\
SUBJ~\cite{subj} & 71.3 & 71.6 & 77.5 & 77.0 \\
CR~\cite{cr}     & 93.8 & 93.8 & 94.7 & 92.9 \\
\midrule
\textbf{Avg} & 82.9 & 83.0 & 86.9 & 86.7 \\
\bottomrule
\end{tabular}}
\end{table}
We later discuss strategies to largely recover this minor loss. Note that context block alignment poses no additional privacy or security risks, sharing the same guarantees as prior KV-cache reuse methods (e.g., RadixCache).

\mypar{(2) De-duplicating multi-turn overlaps reduces prefill cost}
\autoref{fig:example_multi-turn} shows that multi-turn retrievals often return overlapping context blocks across conversation turns.
By deduplicating these blocks and processing only new content together with dialogue history, the amount of contextual data during prefill can be greatly reduced, lowering computation cost.
Beyond whole-block duplication, distinct context blocks often share content at a finer granularity—as underlined in the figure, where overlapping content about Kennedy's death date spans multiple blocks.
Such content-level overlap is especially prevalent when context blocks originate from shared templates or contain standardized sections, as commonly seen in contract analysis, financial filings, and code repositories.

Our MT-RAG trace study quantifies this benefit and shows that de-duplication causes only 1–3\% accuracy degradation, which can be recovered with techniques discussed later.
This minor loss occurs because the LLM can still access the deduplicated content through prior conversation history, preserving quality while avoiding duplicated computation.

\mypar{(3) Context annotation compensates for accuracy perturbation from alignment and de-duplication}  Carefully designed context annotation can largely compensate for the negligible accuracy perturbation introduced by context block alignment and de-duplication.
As shown in~\autoref{fig:order_recovery}, these annotations help the model reconstruct the original ordering relationships within its internal tokens, mitigating accuracy loss.

Evaluations show that context annotations effectively recover accuracy—and in multi-hop reasoning tasks, even improve it beyond the unordered baseline.
On NarrativeQA and MultihopRAG, for instance, accuracy increases by 0.3–3.9\% relative to approximate matching methods (\autoref{subsec:eval-breakdown}), confirming the effectiveness of incorporating context annotations for enhanced reasoning.

Note that the context annotation does not affect the model's instruction-following ability, as it only conveys minimal retrieval metadata without altering the user prompt.

\begin{figure}[!t]
    \centering
    \includegraphics[width=0.85\linewidth]{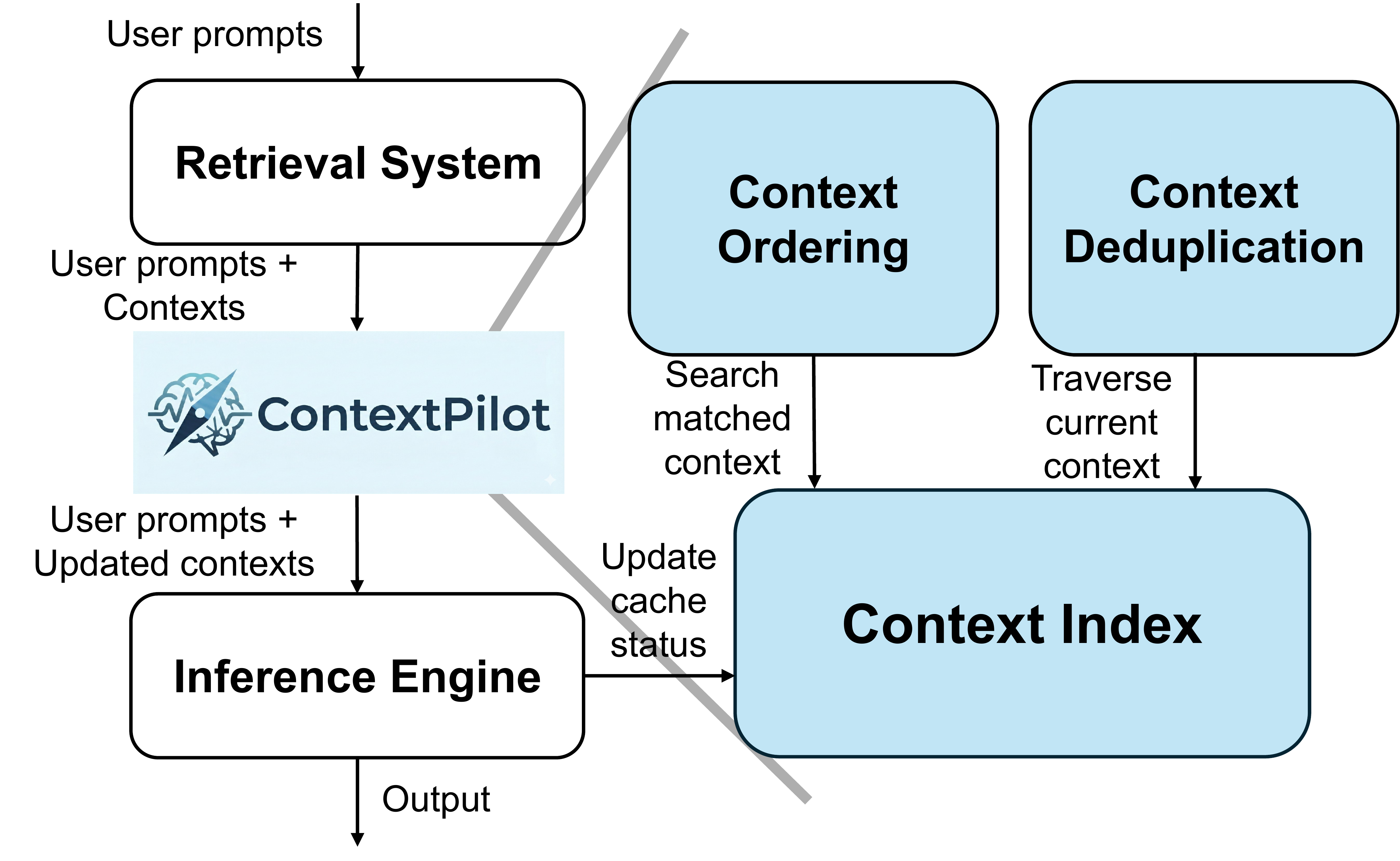}
    \vspace{-0.4cm}
    \caption{System Overview of ContextPilot.}
    \label{fig:system-overview}
    \vspace{-0.7cm}
\end{figure}

\subsection{ContextPilot system overview}

Prior work treats accuracy (e.g., context graphs, agentic memory) and system performance (exact prefix caching) in isolation. \textsc{ContextPilot} uniquely bridges this gap through three contributions: (1)~a context index with a novel distance function that actively aligns documents with the prefix cache to maximize reuse---converting cache misses into hits that prior systems cannot achieve; (2)~succinct context annotations that allow LLMs to recover semantic priority despite aligning; and (3)~multi-turn context traversal that identifies and deduplicates previously memorized documents, reducing prefill overhead especially under model context length constraints. This co-design enables simultaneous gains in efficiency and quality that neither approach achieves alone.

\textsc{ContextPilot} realizes the three design opportunities above, achieving \emph{context reuse with negligible accuracy loss}.
It features a clean, minimal interface compatible with common retrieval modules (e.g., FAISS and ElasticSearch), AI memory layer stores (e.g., Mem0), and inference engines (e.g., SGLang and vLLM), requiring only request ID tracking in the prefix cache of each engine without affecting existing functionality, enabling rapid deployment.
Specifically, \textsc{ContextPilot} takes user prompts and their context blocks (retrieved documents, chunks, or memories), updates the context to enable effective reuse, and then passes the updated context to the inference engine for processing.

\autoref{fig:system-overview} illustrates the key components in \textsc{ContextPilot}: a \emph{context index} that tracks prefix-cache state, a \emph{context alignment mechanism} that aligns and schedules context blocks with the prefix cache for maximum cache hits, and a \emph{context de-duplication mechanism} that removes redundant blocks across multi-turn conversations. The following sections describe each component in detail.

\section{Context Index}\label{sec:indexing}

The context index is designed to: 
(1) efficiently track the inference engine’s prefix-cache status to enable KV-cache reuse; 
(2) support fast lookup of previously stored KV caches via prefix matching, enabling cross-session context reuse when overlaps exist; and 
(3) traverse KV caches in multi-turn conversations to detect duplicated context.

\subsection{Key designs for context index}

\begin{figure}[!t]
    \centering
    \includegraphics[width=\linewidth]{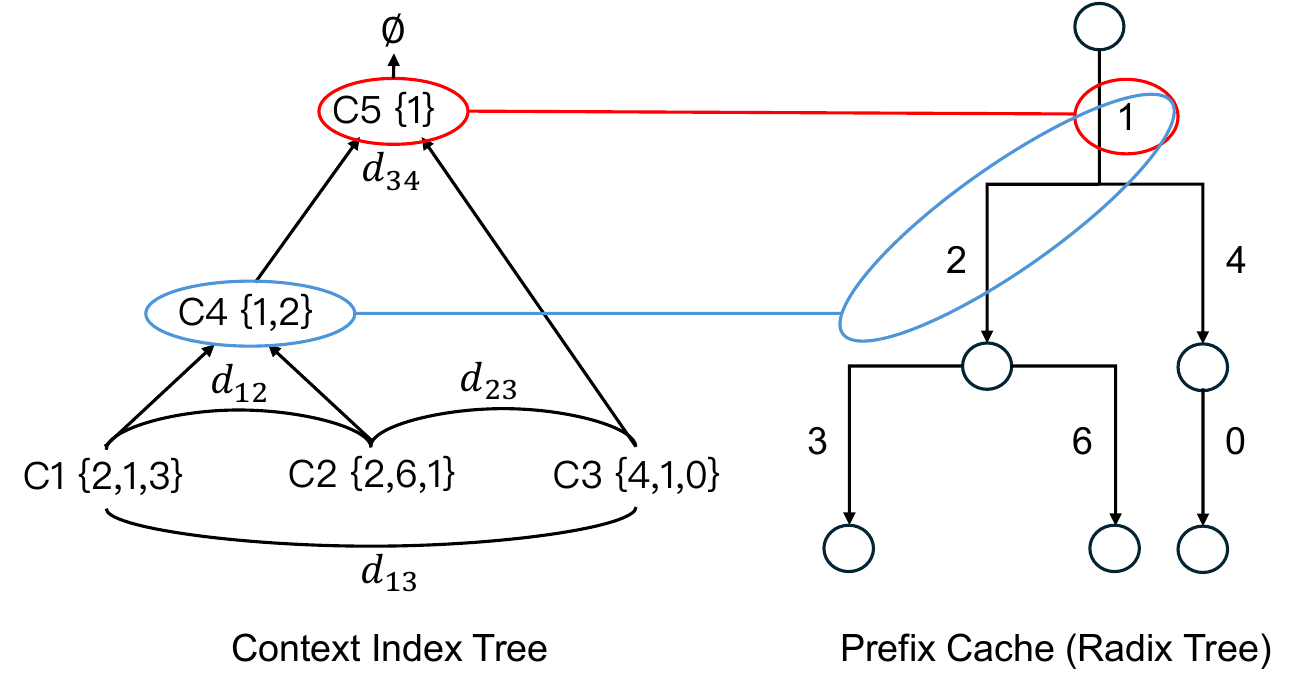}
    \vspace{-0.8cm}
    \caption{Context index construction with prefix-cache semantics.}
    \label{fig:construction}
    \vspace{-2em}
\end{figure}

\autoref{fig:construction} illustrates the structure of the context index with an example. The left panel shows the index tree, and the right panel shows the corresponding prefix-cache status. The index is organized as a tree whose root represents an empty context. Each node corresponds to a prefix stored in the prefix cache and contains child nodes that extend this prefix. Every node maintains four attributes: (1) the context containing context block IDs, (2) the search path from the root to this node, (3) an access frequency counter for cache eviction, and (4) the clustering distance at which the node was created.

\mypar{Index creation}
The index is built via hierarchical clustering based on prefix matching. First, we compute pairwise distances between all contexts using their overlap rate. Next, we iteratively merge the closest pair, creating a virtual node whose context is the sorted intersection representing their shared prefix. Finally, each leaf node records its search path from the root, enabling efficient traversal for both cross-session prefix matching and multi-turn duplicate detection.

As shown in~\autoref{fig:construction}, the process begins with C1 $\{2,1,3\}$, C2 $\{2,6,1\}$, and C3 $\{4,1,0\}$ as leaf nodes. Since C1 and C2 have the smallest distance (sharing $\{1,2\}$), they merge first into a virtual node C4 with context $\{1,2\}$. C3 then merges with C4 to form the root C5 with context $\{1\}$. The resulting tree has C1–C3 as leaves storing their search paths from C5, while C4 and C5 serve as virtual nodes representing shared prefixes for cache reuse.

This construction runs in $O(N^2)$ time, where $N$ is the number of contexts, and is fully parallelizable on CPUs and GPUs. In practice, building the index for 2{,}000 contexts takes 8~s on CPUs and 0.82~s on GPUs. The space complexity is $O(N \cdot K)$, where $K$ is the average number of context blocks per query. Because the index stores only block IDs and metadata rather than full texts, its space overhead is minimal. The complete hierarchical clustering pseudocode is provided in \autoref{alg:cluster-reorder} (\appref{apd:algorithms}).

\mypar{Quantifying the overlapping between contexts}
A key challenge in index construction is quantifying the overlap between contexts. We propose a \emph{context distance function} that satisfies two requirements: (1) it captures the number of shared documents between contexts, and (2) it accounts for their positional alignment, since retrieval systems rank documents by query relevance.

To illustrate the need for this design, consider four contexts: 
A $\{3,5,1,7\}$, B $\{2,6,3,5\}$, C $\{3,5,8,9\}$, and D $\{2,6,4,0\}$. 
A naive overlap-only metric assigns identical distances (0.5) to pairs A–B, B–C, and B–D because each shares two documents. 
However, B and D share $\{2,6\}$ at positions 1–2, while A and B share $\{3,5\}$ at different positions. 
Our distance function (\autoref{eq:distance_metric}) assigns a smaller distance to B–D, as their overlaps occur in similar positions, reflecting both overlap magnitude and positional alignment. 
Such patterns cannot be captured by conventional distance measures like cosine, L1, or L2 similarity, which ignore positional structure. More formally, our distance function is defined as:
\begin{equation}
\label{eq:distance_metric}
\small
d_{ij} = 1 - \frac{|S_{ij}|}{\max(|C_i|, |C_j|)} + \alpha \cdot \frac{\sum_{k \in S_{ij}} |p_i(k) - p_j(k)|}{|S_{ij}|}
\end{equation}
where $S_{ij}$ denotes the set of shared documents, $p_i(k)$ is the position of document $k$ in context $i$, and $\alpha \in [0.001, 0.01]$ ensures overlap count remains the dominant factor while incorporating positional alignment.

\mypar{Index update}
The context index stays synchronized with the inference engine’s prefix cache through lightweight request ID tracking. Each leaf node is associated with a request ID maintained by the engine. When the engine evicts cached entries, it sends the corresponding request IDs to ContextPilot, which looks up the affected nodes via a request-to-node mapping and removes them. Empty parent nodes are recursively pruned to keep the tree compact. The overall update cost is $O(h)$, where $h$ is the tree height, requiring only a single traversal per eviction.

\subsection{Key operations with context index}

The context index provides two key operations:

\mypar{Context search} 
\textsc{ContextPilot} frequently searches for previously stored contexts based on the current one to enable reuse. 
The index search algorithm (\autoref{alg:search}, \appref{apd:algorithms}) efficiently locates matching contexts by greedily descending from the root, selecting at each level the child with the minimum distance while recording positions to form a search path.
The search stops upon reaching a leaf or when all children are equidistant, indicating the longest shared prefix.
Updates are localized and efficient: matching an internal node appends the new context as a child ($O(1)$), while matching a leaf creates a new internal node with their intersection ($O(|C|)$).
Unlike K-Means re-clustering or HNSW graph rebuilding, these updates require no tree restructuring, enabling dynamic index maintenance with minimal overhead.

For example, given context C6 $\{2,1,4\}$, we search the index in \autoref{fig:construction}. 
C6 first compares with the root’s child C5 and finds a shared prefix $\{1\}$, descending to C5 and recording its position [0]. 
At C5, C6 shares $\{1,2\}$ with C4 but only $\{1\}$ with C3, so it selects C4 and appends another [0], yielding [0,0]. 
At C4’s children C1 $\{1,2,3\}$ and C2 $\{1,2,6\}$, all have equal distance, so the search stops and identifies C4 as the best match with path [0,0]. 
C6 is then inserted into C4’s children list at position 2, forming the final search path [0,0,2]. 

Search complexity scales with tree height. For contexts with common prefixes, $h = O(\log n)$ yields $O(|C| \cdot \log n)$ complexity, where $n$ denotes the number of stored contexts. Empirically, search takes approximately 0.068~ms per request (\appref{apd:overhead}), negligible compared to prefill latency.

\mypar{Context traversal}
In multi-turn conversations, \textsc{ContextPilot} updates node context lengths by traversing the index using the stored search path.
Starting from the root, it sequentially follows indices along the path until reaching the target node, then performs the update.
Traversal costs $O(h)$ and is subsumed by the search overhead above.

\section{Context Alignment} \label{sec:ordering}
\begin{figure*}[!t]
    \centering
    \includegraphics[width=\linewidth]{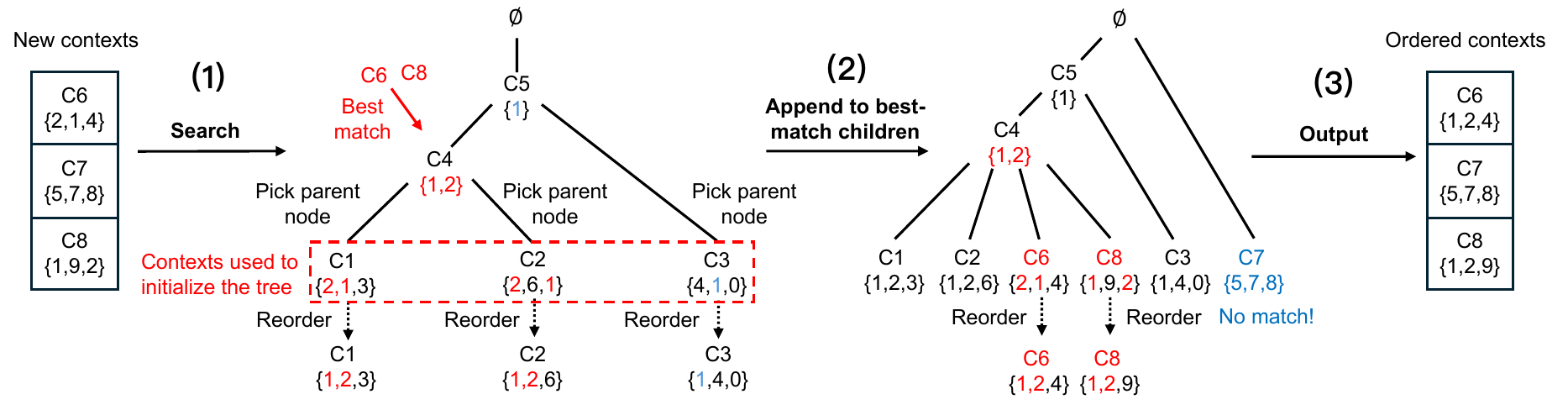}
    \vspace{-0.8cm}
    \caption{Example for aligning context with prefix cache: (1) find best-matching nodes, (2) align by shared prefix and append as a child, and (3) output aligned context.}
    \vspace{-0.5cm}
    \label{fig:intra-order}
\end{figure*}
The context alignment mechanism aims to: 
(1) align incoming contexts with the current prefix cache to maximize KV-cache reuse;
(2) schedule the aligned contexts to the inference engine with awareness of cache generation and eviction policies to enhance hit ratio; and 
(3) insert context annotations that recover pre-alignment semantics and maintain accuracy.

\subsection{Context alignment algorithm}

Formally, the context alignment algorithm (\autoref{alg:context-ordering}, \appref{apd:algorithms}) takes a batch of requests with their context blocks as input, aligns them with the prefix cache based on prefix matches from the context index, and returns aligned contexts with maximized shared prefixes.

As illustrated in~\autoref{fig:intra-order}, we begin with initialization contexts C1 $\{2,1,3\}$, C2 $\{2,6,1\}$, and C3 $\{4,1,0\}$, followed by new contexts C6 $\{2,1,4\}$, C7 $\{5,7,8\}$, and C8 $\{1,2,9\}$. 
Initialization contexts inherit prefixes from their parent nodes (C1, C2 from C4 with $\{1,2\}$; C3 from C5 with $\{1\}$), while new contexts search the index (C6 and C8 match C4 and inherit $\{1,2\}$). 
Each context then concatenates its matched prefix with remaining documents in their original order, producing C1 $\rightarrow$ $\{1,2,3\}$, C2 $\rightarrow$ $\{1,2,6\}$, C6 $\rightarrow$ $\{1,2,4\}$, and C8 $\rightarrow$ $\{1,2,9\}$. 
Unmatched contexts (e.g., C7) remain unchanged and form standalone branches. 
This strategy ensures overlapping contexts share common prefixes while preserving the ranking of non-shared documents.

The algorithm is invoked whenever \textsc{ContextPilot} processes a new request. 
It runs in $O(|C|\cdot \log n)$ time, where $n$ is the number of stored contexts, taking approximately 0.047~ms per request (\appref{apd:overhead})—negligible compared to prefill.

\subsection{Scheduling requests with aligned contexts}

After aligning contexts, \textsc{ContextPilot} must schedule their execution to match the inference engine’s KV-cache generation and eviction policies; otherwise, cache reuse becomes ineffective. 
We therefore design a scheduling algorithm that: 
(1) reuses the search paths obtained during context alignment to avoid redundant tree lookups; 
(2) groups contexts by the first element of their search path, naturally separating cache regions; and 
(3) sorts contexts within each group by path length in descending order, ensuring longer prefix matches execute before shorter ones.

\begin{figure}[t!]
    \centering
    \includegraphics[width=\linewidth]{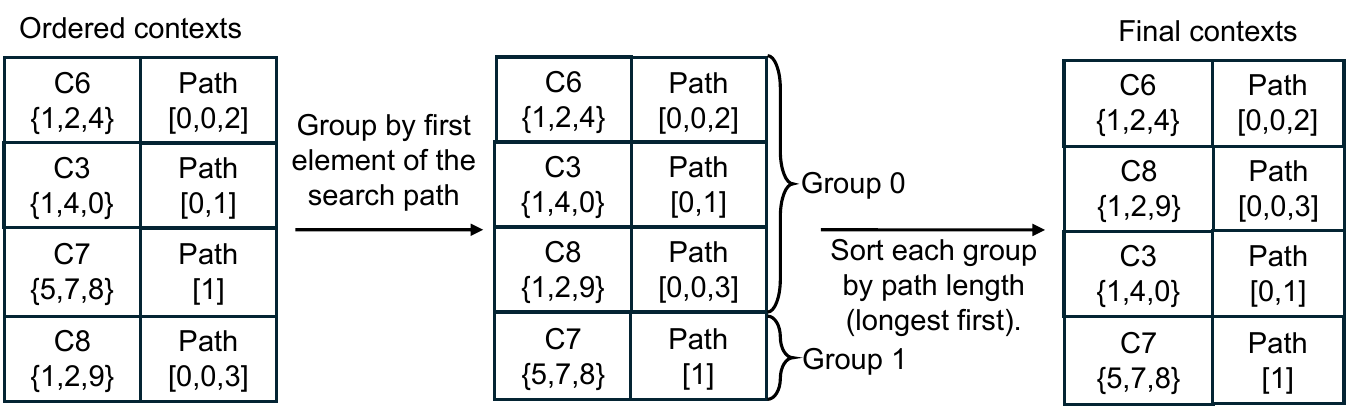}
    \vspace{-0.8cm}
    \caption{Example of scheduling requests with aligned contexts.}
    \vspace{-0.4cm}
    \label{fig:inter-reorder}
\end{figure}

\autoref{fig:inter-reorder} illustrates this process. 
In the baseline order C6, C3, C7, C8, limited cache capacity allows only one context: 
C6 caches $\{1,2,4\}$, but C3 reuses only $\{1\}$ and evicts $\{2,4\}$. 
C7 causes a full miss, caching $\{5,7,8\}$ and evicting all previous entries, which then forces another miss for C8 despite its shared prefix $\{1,2\}$ with C6. 
This inefficiency arises because contexts with shared prefixes are not executed consecutively.

Our scheduler rearranges the execution to C6, C8, C3, C7, grouping prefix-sharing contexts together. 
C6 first caches $\{1,2,4\}$, then C8 immediately reuses $\{1,2\}$ before eviction. 
C3 and C7 run afterward without disrupting this reuse, maximizing cache hit ratio.

Our scheduler performs $O(N)$ grouping by root-prefix path and \({O}(N\log N)\) in-group sorting over \(N\) contexts, with negligible real-time overhead. In contrast, existing indexing methods such as RAGCache and SGLang's LPM use a \textbf{global prefix selection that rescans a radix tree with \(M\) nodes at each decision point}, yielding \(O(N\log M) + O(N\log N)\) overall as cache state evolves. By draining groups sequentially, our method \textbf{avoids repeated tree searches}, better preserves reuse under tight KV budgets, and keeps complexity independent of \(M\). The full scheduling pseudocode is given in \autoref{alg:grouping} (\appref{apd:algorithms}).

\subsection{Context annotation for context alignment}

\mypar{Why aligning is safe}
As shown in \autoref{subsec:design-opp}, our reproduction of the DEmO study~\cite{demo} confirms that modern LLMs are substantially less sensitive to input ordering (\autoref{tab:demo-reproduce}), explaining why aligning for cache efficiency introduces only minor accuracy perturbation and making lightweight correction mechanisms sufficient.

\mypar{Why annotations still help}
Despite the reduced sensitivity, aligning can still cause minor accuracy perturbation on some datasets (e.g., $-$1.1\% on QASPER). Annotations mitigate this by reducing the model's reliance on positional signals to infer relevance.
On multi-hop tasks where chaining evidence across context blocks benefits from explicit guidance, annotations not only recover lost accuracy but actively \emph{improve} it beyond the no-alignment baseline (e.g., +4.0\% F1 on MultihopRAG with Qwen3-32B; see \appref{apd:accuracy-breakdown}).
Gains are consistent across model scales: Qwen3-4B gains +1.4\% on MultihopRAG and +1.3\% on NarrativeQA, while Qwen3-32B gains +4.0\% and +1.2\% respectively.
Attention map analysis (\appref{apd:attn}) confirms that annotations reshape internal attention, aligning it with semantic rather than positional priority.

\mypar{Annotation mechanism}
We provide the LLM with succinct annotations indicating the original relevance ranking of context blocks. Aligning contexts alters this ranking, which encodes document relevance critical for answer quality. 
Consider context C6, where the retriever returns documents in order $\{2,1,4\}$. 
The baseline prompt is:
\begin{center}
\small
[system prompt] $\rightarrow$ [CB\_2] $\rightarrow$ [CB\_1] $\rightarrow$ [CB\_4] $\rightarrow$ [question]
\end{center}
After aligning to $\{1,2,4\}$ for cache efficiency, we append an \emph{order annotation} before the question:
\begin{center}
\small
[system prompt] $\rightarrow$ [CB\_1] $\rightarrow$ [CB\_2] $\rightarrow$ [CB\_4] $\rightarrow$ [order annotation] $\rightarrow$ [question]
\end{center}
The annotation explicitly specifies the original relevance priority:
\begin{center}
\textit{``Please read the context in the following priority order: [CB\_2] $>$ [CB\_1] $>$ [CB\_4] and answer the question.''}
\end{center}

This short instruction adds negligible token overhead during prefill yet effectively preserves the model’s ability to attend to documents by their original relevance ranking.
As a result, \textsc{ContextPilot} achieves aggressive cache optimization with minor accuracy perturbation (${\leq}$1\% on most datasets), and often improved accuracy on multi-hop reasoning tasks.

\section{Context Deduplication} \label{sec:dedup}

The context de-duplication mechanism has two goals:
(1) eliminate redundant content—both entire context blocks repeated across turns and shared content across distinct context blocks—whose KV caches are already stored, thereby minimizing redundant prefill computation; and
(2) provide annotations informing the LLM which content has been de-duplicated and where the corresponding information resides in the earlier context.

\autoref{alg:dedup} (\appref{apd:algorithms}) formalizes the full de-duplication procedure, which operates at two levels. The algorithm runs in $O(|C|)$ time, taking approximately 0.6~ms per request (\appref{apd:overhead})—negligible compared to prefill.

\mypar{Context-block-level de-duplication} The context index maintains a per-conversation record of all context blocks processed in prior turns, populated when the first turn's context is aligned and inserted into the index (\autoref{sec:ordering}). Given a new context, the algorithm looks up these previously indexed blocks, identifies exact matches, and generates a location annotation for each duplicate. The current turn's blocks are then registered for future comparisons.

We illustrate with an example.
Consider a session where user C6 initially retrieves context $\{1,2,4\}$ in the first turn. During context alignment (\autoref{sec:ordering}), these blocks are inserted into the context index and recorded in C6's conversation history.
In the second turn, a new query yields $\{1,5,2\}$.
The algorithm queries the index's conversation record, identifies $\{1,2\}$ as already cached from the first turn, and replaces them with location annotations, leaving only the novel block $\{5\}$ to be fully processed.

\mypar{Content-level de-duplication}
Context-block-level de-duplication handles exact matches across turns, but distinct context blocks often share overlapping content—as illustrated in \autoref{fig:example_multi-turn}, where information about Kennedy's death date appears across multiple context blocks.
Inspired by content-defined chunking in deduplication storage systems~\cite{lbfs}, we split each novel context block into variable-length sub-blocks at boundaries where $\textsc{Hash}(\ell) \bmod M = 0$ for a text line $\ell$.
Unlike fixed-size chunking, where a single insertion or deletion shifts all subsequent boundaries and prevents hash matches, content-defined boundaries are determined solely by local content, ensuring that identical text always produces the same sub-blocks regardless of its offset within different context blocks.

Each sub-block is hashed. Any sub-block matching a hash from a different context block is replaced with a location annotation pointing to the first occurrence. After deduplication, the context index is updated to reflect these changes for the request.

\mypar{Context annotation for de-duplicated context blocks} Simply removing duplicates can degrade answer quality.
To maintain quality, we insert \emph{location annotations} (e.g., \textit{``Please refer to [CB\_1] in the previous conversation’’}) that direct the LLM to corresponding context blocks in the conversation history, guiding the LLM to prior context without repeating prefill. For C6’s second turn, the prompt changes from:
\begin{center}
\small
[first turn context] $\rightarrow$ [first turn Q\&A] $\rightarrow$ [CB\_1] $\rightarrow$ [CB\_5] $\rightarrow$ [CB\_2] $\rightarrow$ [second turn question]
\end{center}
\vspace{-0.3cm}
to:
\vspace{-0.1cm}
\begin{center}
\small
[first turn context] $\rightarrow$ [first turn Q\&A] $\rightarrow$ [annotation\_1] $\rightarrow$ [CB\_5] $\rightarrow$ [annotation\_2] $\rightarrow$ [second turn question]
\end{center}

\section{Evaluation}

Our evaluation of \textsc{ContextPilot} shows: (1)~\textsc{ContextPilot} improves prefill throughput by up to 3.1$\times$ and accuracy by up to 4.0\% over numerous state-of-the-art systems and methods across multi-turn, multi-session, and hybrid RAG workloads, (2)~It outperforms strong baselines by 1.5-3$\times$ in throughput and 1.9-3.7\% in accuracy in emerging agentic AI applications, including real-world agent deployments, (3)~Each component (context alignment, de-duplication, and annotation design) yields clear gains and robustness with negligible overhead, (4)~Benefits scale with longer contexts and larger retrieval sizes, and (5)~gains extend to edge devices, achieving ${\sim}2.4\times$ latency reduction on consumer hardware without GPU servers.

\begin{table*}[!ht]
  \small
  \setlength{\tabcolsep}{3.4pt}
  \centering
  \caption{Multi-session RAG results: F1 (\%) and prefill throughput for four methods across three models on three datasets.}
  \label{tab:multi-session}
  \resizebox{\linewidth}{!}{%
  \begin{tabular}{ll cc cc cc cc}
    \toprule
    & & \multicolumn{2}{c}{LMCache} & \multicolumn{2}{c}{CacheBlend} & \multicolumn{2}{c}{Radix Cache} & \multicolumn{2}{c}{\textbf{ContextPilot (Ours)}} \\
    \cmidrule(lr){3-4}\cmidrule(lr){5-6}\cmidrule(lr){7-8}\cmidrule(lr){9-10}
    \textbf{Dataset} & \textbf{Model} & F1 & Prefill Throughput & F1 & Prefill Throughput & F1 & Prefill Throughput & F1 & Prefill Throughput \\
    \midrule
   \multirow{3}{*}{MultihopRAG}
      & Qwen3-4B-Instruct-2507  & 35.2 & 34710.9 & 34.8 & 85405.3 & 35.2 & 58990.1 & \textbf{36.6} & \textbf{106799.5} \\
      & Qwen3-32B               & 60.4 & 14708.6 & 50.1 & 36128.6 & 60.4 & 17682.6 & \textbf{64.4} & \textbf{36296.1} \\
      & Llama3.3-70B-Instruct   & \textbf{62.9} & 11596.4 & 54.9 & 14134.2 & \textbf{62.9} & 14777.1 & \textbf{62.9} & \textbf{30046.7} \\
    \midrule
    \multirow{3}{*}{NarrativeQA}
      & Qwen3-4B-Instruct-2507  & 16.0 & 39276.4 & 11.3 & 42819.5 & 16.0 & 39492.4 & \textbf{17.3} & \textbf{57681.3} \\
      & Qwen3-32B               & 28.4 & 15514.0 & 19.8 & 16913.2 & 28.4 & 15598.8 & \textbf{29.6} & \textbf{22780.4} \\
      & Llama3.3-70B-Instruct   & 37.8 & 12575.7 & 31.3 & 13710.5 & 37.8 & 12644.9 & \textbf{38.4} & \textbf{18468.8}  \\
    \midrule
    \multirow{3}{*}{QASPER}
      & Qwen3-4B-Instruct-2507  & \textbf{27.9} & 29349.2 & 21.9 & 36273.5 & \textbf{27.9} & 33034.6 & 26.8 & \textbf{46619.9} \\
      & Qwen3-32B               & \textbf{36.0} & 15568.4 & 29.3 & 20238.9 & \textbf{36.0} & 17523.0 & 34.9 & \textbf{24733.4} \\
      & Llama3.3-70B-Instruct   & \textbf{33.8} & 12000.1 & 27.9 & 14829.7 & \textbf{33.8} & 13507.3 & \textbf{33.8} & \textbf{19061.0} \\
    \bottomrule
  \end{tabular}}
  \vspace{-1.5em}
\end{table*}

\mypar{Evaluation setup} Our \textsc{ContextPilot} implementation supports SGLang~0.4.6 and vLLM~0.10.0, requiring only request ID tracking in each engine's prefix cache without affecting existing functionality, making the changes easy to upstream and merge.

We compare \textsc{ContextPilot} against the following baselines:  
(i)~\textsc{LMCache}~(version 0.3.8), representing the state of the art in prompt caching;  
(ii)~\textsc{CacheBlend}, the state of the art in KV-cache matching, integrated with LMCache; and  
(iii)~\textsc{RadixCache}, based on SGLang’s implementation with a Longest-Prefix-Match scheduling policy.

We omit additional baselines that achieve comparable performance to those above.  
(iv)~\textsc{HiCache}~\cite{xie2025stratahierarchicalcontextcaching} extends RadixCache by expanding prefix caches to lower-tier memory; since it directly builds on RadixCache, we compare against RadixCache instead.  
(v)~\textsc{RAGCache} adopts a similar radix-tree structure at document granularity and shows comparable performance to RadixCache. It is not open-sourced and thus excluded from our evaluation.

Our evaluation is conducted on two GPU clusters: 
(1) a 16×~H100 GPU cluster and (2) a 12×~A6000 GPU cluster. For all baselines, we tune system parameters for optimal performance and accuracy, aligning our configurations with the best results reported in their respective papers. We set $\alpha = 0.001$ for the distance metric in \autoref{eq:distance_metric} across all experiments.

% \mypar{Evaluation metrics} We report widely adopted metrics such as F1-overlapping and LLM-as-a-Judge, matching SOTA LLM reasoning studies. 

\subsection{Performance in Retrieval-Augmented Generation}\label{subsec:eval-app}

We evaluate on four RAG datasets: QASPER~\cite{qasper}, MultihopRAG~\cite{mulhoprag}, NarrativeQA~\cite{narrativeqa}, and MT-RAG~\cite{mtrag}. 
For QASPER, MultihopRAG, and NarrativeQA, we use a chunk size of 1024 following~\cite{bhat2025rethinkingchunksizelongdocument}, while MT-RAG performs document-level retrieval without chunking. 
We use \texttt{gte-Qwen2-7B-Instruct} as the embedding model, FAISS for similarity search on MultihopRAG and NarrativeQA, and BM25 for QASPER and MT-RAG. 
This setup demonstrates \textsc{ContextPilot}'s effectiveness across diverse retrieval paradigms.

We further test \textsc{ContextPilot} under three RAG workloads:
(1) multi-session, where independent users query concurrently;
(2) multi-turn, for extended single-user dialogues; and
(3) combined multi-session and multi-turn, representing production-scale conversational systems.
For multi-session experiments, \textsc{ContextPilot} operates in \emph{offline} mode: all contexts are pre-fetched and the context index is built before large-batch inference begins.
For multi-turn and Mem0 experiments, \textsc{ContextPilot} operates in \emph{online} mode with cold-start: the context index is incrementally built and updated as each turn arrives.

\mypar{Multi-session RAG} 
We evaluate multi-session RAG on QASPER, MultihopRAG, and NarrativeQA using three models—Qwen3-4B-Instruct-2507, Qwen3-32B, and Llama3.3-70B-Instruct—on H100 GPUs with top-$k{=}15$.  

\autoref{tab:multi-session} reports F1 scores and prefill throughput.
\textsc{ContextPilot} delivers up to 3.08×, 2.05×, and 2.13× speedups over LMCache, RadixCache, and CacheBlend on MultihopRAG, and 1.3–1.6× gains on NarrativeQA and QASPER, by aligning contexts with the prefix cache to maximize overlap.
In contrast, LMCache and RadixCache depend on exact prefix matching, causing recomputation even for overlapping content, while LMCache also incurs high CPU offloading costs for long contexts.
\textsc{ContextPilot} also maintains or improves accuracy via order annotations (e.g., 60.4$\to$64.4 on MultihopRAG with Qwen3-32B), while CacheBlend degrades sharply (F1 drops to 11.3 on NarrativeQA with Qwen3-4B) due to approximate KV matching disrupting coherence.

\mypar{Effectiveness with ever larger MoE models}
We further evaluate \emph{DeepSeek-R1 (671B)} on a GPU cluster with 32 H20 GPUs, provided by a potential industry adopter. On MultihopRAG, \textsc{ContextPilot} increases the cache hit ratio from 5\% to 60\%; on NarrativeQA, it raises the hit ratio from 6\% to 38\%. These improvements translate into 1.81x and 1.52x higher prefill throughput on 16xH20, respectively. Scaling to 32xH20 yields similar speedups, confirming that our approach generalizes to larger reasoning models and multi-node deployments. We provide a more detailed analysis of the DeepSeek-R1 results in~\appref{apd:deepseek-r1}.
%%%%%%%%%%%%%%%%%%%%%%%%table 2%%%%%%%%%%%%%%%%%%%%%%%%
% \begin{table}[!t]
%     \centering
%     % Define a new column type 'C' that is centered and wraps text.
%     % The #1 is a placeholder for the width (e.g., C{2.5cm}).
%     \newcolumntype{C}[1]{>{\centering\arraybackslash}p{#1}}

%     \caption{Acc (\%) and time-to-first-token (TTFT)(s) on MT-RAG for four methods across three models. X = not supported.}
%     \label{tab:one-ds-three-models-transposed}
    
%     \resizebox{0.8\linewidth}{!}{
%     \begin{tabular}{ll C{1.2cm} C{1.2cm} C{1.2cm}}
%         \toprule
%         \textbf{Method} & \textbf{Metric} &Qwen3-4B-Instruct & Llama3.1-8B-Instruct & Qwen3-30B-A3B \\
%         \midrule
%         \multirow{2}{*}{LMCache}  
%         & Acc  & 62.56 & \textbf{68.46} & 75.12  \\
%         & TTFT & 0.76 & 1.04  & 1.08  \\
%         \cmidrule(lr){2-5}
%         \multirow{2}{*}{CacheBlend}
%         & Acc  & 50.33 & 56.52 & X \\
%         & TTFT & 0.30 & 0.48 & X \\
%         \cmidrule(lr){2-5}
%         \multirow{2}{*}{Radix Cache}
%         & Acc  & 62.56  & \textbf{68.46}  & 75.12  \\
%         & TTFT &  0.44 &  0.49 & 0.61  \\
%         \cmidrule(lr){2-5}
%         \multirow{2}{*}{\textbf{Ours}}        
%         & Acc  &  \textbf{64.27} & 68.12  & \textbf{75.81}  \\
%         & TTFT & \textbf{0.22} &  \textbf{0.31} & \textbf{0.35} \\
%         \bottomrule
%     \end{tabular}
%     }
% \end{table}

\begin{table*}[!t]
    \centering
    \caption{Performance metrics across different tasks: (a) MT-RAG results, (b) Hybrid RAG sessions, (c) Context index construction latency.}
    \label{tab:combined-tables}
    \vspace{0.1cm}
    
    % First minipage - Table (a)
    \begin{minipage}[b]{0.32\textwidth}
        \centering
        \newcolumntype{C}[1]{>{\centering\arraybackslash}p{#1}}
        \resizebox{\linewidth}{!}{
        \begin{tabular}{ll C{1.2cm} C{1.2cm} C{1.2cm}}
            \toprule
            \textbf{Method} & \textbf{Metric} & Qwen3-4B & Llama3.1-8B & Qwen3-30B \\
            \midrule
            \multirow{2}{*}{LMCache}  
            & Acc.  & 62.56 & \textbf{68.46} & 75.12  \\
            & TTFT & 0.76 & 1.04  & 1.08  \\
            \cmidrule(lr){2-5}
            \multirow{2}{*}{CacheBlend}
            & Acc.  & 50.33 & 56.52 & X \\
            & TTFT & 0.30 & 0.48 & X \\
            \cmidrule(lr){2-5}
            \multirow{2}{*}{RadixCache}
            & Acc.  & 62.56  & \textbf{68.46}  & 75.12  \\
            & TTFT &  0.44 &  0.49 & 0.61  \\
            \cmidrule(lr){2-5}
            \multirow{2}{*}{\textbf{ContextPilot}}        
            & Acc.  &  \textbf{64.27} & 68.12  & \textbf{75.81}  \\
            & TTFT & \textbf{0.22} &  \textbf{0.31} & \textbf{0.35} \\
            \bottomrule
        \end{tabular}
        }
        % \vspace{0.1cm}
        \subcaption{Accuracy (\%) and time-to-first-token (TTFT)(s) on MT-RAG for four methods across three models. X = not supported.}
        \label{tab:one-ds-three-models-transposed}
    \end{minipage}%
    \hfill
    % Second minipage - Table (b)
    \begin{minipage}[b]{0.32\textwidth}
        \centering
        \resizebox{\linewidth}{!}{%
        \begin{tabular}{lccccc}
            \toprule
            & \multicolumn{5}{c}{\textbf{\# Sessions}} \\
            \cmidrule(lr){2-6}
            \textbf{Method} & \textbf{2} & \textbf{4} & \textbf{8} & \textbf{16} & \textbf{32} \\
            \midrule
            LMCache      & 0.81 & 0.87 & 0.94 & 1.19 & 1.72 \\
            CacheBlend   & 0.40 & 0.42 & 0.45 & 0.54 & 0.78 \\
            RadixCache   & 0.46 & 0.49 & 0.53 & 0.67 & 0.97 \\
            \textbf{ContextPilot} & \textbf{0.24} & \textbf{0.29} & \textbf{0.34} & \textbf{0.41} & \textbf{0.65} \\
            \bottomrule
        \end{tabular}}
        \vspace{0.175cm}
        % \subcaption{TTFT (s) for Qwen3-4B-Instruct-2507 in hybrid RAG across different session counts.}
        \subcaption{Time-to-first-token (seconds) performance in seconds for Qwen3-4B-Instruct-2507 model evaluated under hybrid RAG workloads with varying concurrent session counts ranging from 2 to 32 sessions.}
        \label{tab:ttft-sessions}
    \end{minipage}%
    \hfill
    % Third minipage - Table (c)
    \begin{minipage}[b]{0.32\textwidth}
        \centering
        \small
        \resizebox{\linewidth}{!}{%
        \begin{tabular}{lcccccc}
            \toprule
            & \multicolumn{6}{c}{\textbf{\# Contexts} ($N_{\text{ctx}}$)} \\
            \cmidrule(lr){2-7}
            \textbf{$k$} & \textbf{128} & \textbf{512} & \textbf{4k} & \textbf{8k} & \textbf{12k} & \textbf{100k} \\
            \midrule
             3  & 0.64 & 0.65  & 1.51 & 3.54 & 7.48 & 687.64\\
             5  & 0.66 & 0.66  & 1.55 & 3.58 & 7.55 & 688.21\\
             10 & 0.67 & 0.68  & 1.59 & 3.63 & 7.63 & 689.07\\
             15 & 0.69 & 0.69  & 1.62 & 3.67 & 7.69 & 687.95\\
             20 & 0.71 & 0.72  & 1.66 & 3.72 & 7.78 & 690.12\\
            \bottomrule
        \end{tabular}
        }
        \vspace{0.02cm}
        \subcaption{Context index construction latency (seconds) as a function of the number of contexts $N_{\text{ctx}}$ and top-$k$.
  Columns labeled \textbf{128}–\textbf{100k} denote the total contexts inserted at build time ($N_{\text{ctx}}$; \textit{1k} $=$ 1{,}000).}
        \label{tab:gpu-latency-multik}
    \end{minipage}
    \vspace{-1em}
\end{table*}

\mypar{Multi-turn RAG} 
We evaluate multi-turn RAG on the MT-RAG dataset using Qwen3-4B-Instruct-2507, Llama3.1-8B-Instruct, and Qwen3-30B-A3B-Thinking-2507 on a single H100 GPU. 
Models with longer context windows are used to handle the growing conversation history. 
Answer accuracy is measured via the LLM-as-a-judge method from RADBench~\cite{kuo2025radbenchevaluatinglargelanguage} with GPT-5, as recommended by MT-RAG.

\autoref{tab:one-ds-three-models-transposed} reports accuracy and time-to-first-token (TTFT).
\textsc{ContextPilot} cuts TTFT by removing redundant document processing across turns through context de-duplication.
It achieves 3.45×, 3.35×, and 3.09× speedups over LMCache on Qwen3-4B, Llama3.1-8B, and Qwen3-30B, respectively, and up to 2.00× over RadixCache and 1.55× over CacheBlend.
\textsc{ContextPilot} also preserves accuracy via location annotations that direct models to previously seen documents (e.g., 62.56\%$\to$64.27\% on Qwen3-4B), while CacheBlend drops to 50.33\% due to approximate KV matching disrupting multi-turn coherence.

%%%%%%%%%%%%%%%%%%%%%%%%table 3%%%%%%%%%%%%%%%%%%%%%%%%
% \begin{table}[!t]
%     \centering
%     \caption{TTFT (s) for Qwen3-4B-Instruct-2507 in hybrid RAG.}
%     \label{tab:ttft-sessions}
%     \resizebox{0.80 \linewidth}{!}{%
%     \begin{tabular}{lccccc}
%         \toprule
%         & \multicolumn{5}{c}{\textbf{\# Sessions}} \\
%         \cmidrule(lr){2-6}
%         \textbf{Method} & \textbf{2} & \textbf{4} & \textbf{8} & \textbf{16} & \textbf{32} \\
%         \midrule
%         LMCache      & 0.81 & 0.87 & 0.94 & 1.19 & 1.72 \\
%         CacheBlend   & 0.40 & 0.42 & 0.45 & 0.54 & 0.78 \\
%         RadixCache   & 0.46 & 0.49 & 0.53 & 0.67 & 0.97 \\
%         \textbf{Ours} & \textbf{0.24} & \textbf{0.29} & \textbf{0.34} & \textbf{0.41} & \textbf{0.65} \\
%         \bottomrule
%     \end{tabular}}
% \end{table}

\mypar{Multi-session, multi-turn RAG} 
We evaluate the combined multi-session and multi-turn scenario under real-world deployment using Qwen3-4B-Instruct-2507 on H100 GPUs, varying concurrency from 2 to 32 sessions.

\autoref{tab:ttft-sessions} shows that \textsc{ContextPilot} achieves the lowest TTFT at all concurrency levels by aligning contexts with the prefix cache.
At 2 sessions, it delivers 3.38×, 1.92×, and 1.67× speedups over LMCache, RadixCache, and CacheBlend, respectively;
at 32 sessions, the gains remain substantial at 2.65×, 1.49×, and 1.20×.
These improvements confirm that context alignment maintains prefix overlap as concurrency scales.

\subsection{Performance in Agentic Applications}

We study three representative agentic workloads that are increasingly common in production---multi-agent reasoning, AI memory, and real-world agent deployment---all of which repeatedly surface long contexts across turns and sessions, making them natural candidates for context reuse.

\mypar{Multi-agent reasoning} We evaluate \textsc{ContextPilot} with \emph{Chain-of-Agent (CoA)}~\cite{zhang2024chain}, where worker agents handle document segments and a manager aggregates results.
\textsc{ContextPilot} enhances CoA through agent-aware routing: in multi-session settings, recurring documents are routed to the agent that previously processed them for KV-cache reuse; in multi-turn conversations, repeated documents are deduplicated with location annotations directing agents to prior content.
We deploy three CoA configurations on MultihopRAG, each with 15 agents using Llama3.1-8B, Llama3.2-3B, or Qwen3-4B-Instruct ($k{=}15$).
With Qwen3-4B, accuracy increases from \textbf{48.3\%} to \textbf{50.2\%} and throughput by \textbf{1.8×}; with Llama3.1-8B, accuracy rises from \textbf{50.7\%} to \textbf{54.4\%} with a \textbf{2.1×} speedup.

\mypar{Agentic memory systems}
We evaluate \textsc{ContextPilot} with Mem0~\cite{mem0}, a popular AI memory system that repeatedly retrieves user-specific memories as context blocks, creating substantial cross-request overlap.
Using Qwen3-4B on LoCoMo~\cite{locomo} with GPT-4.1 as judge, \textsc{ContextPilot} indexes and aligns fetched memories with the prefix cache in online mode.
At $k{=}100$, it reduces TTFT from 0.101\,s to 0.055\,s (\textbf{1.83$\times$} speedup) with a minor accuracy trade-off (0.437$\to$0.420).
At $k{=}20$, although LoCoMo conversations are relatively short (${\sim}$26K tokens across all turns on average), \textsc{ContextPilot} still improves both TTFT (0.038\,s$\to$0.031\,s, 1.23$\times$) and accuracy (0.440$\to$0.460), as alignment places more relevant memories earlier.

\mypar{Real-world agent deployment}
To evaluate \textsc{ContextPilot} in an end-to-end agent pipeline, we integrate it with OpenClaw served via SGLang~0.5.9 on a single RTX~5090.
The OpenClaw agent issues requests through the \textsc{ContextPilot} proxy, which aligns prompts with the prefix cache, deduplicates overlapping context both across turns and within context blocks, and forwards optimized prompts to the inference engine (\autoref{fig:openclaw} in \appref{apd:openclaw}).
We evaluate on the \texttt{claw-tasks} benchmark using Qwen3-4B-Instruct-2507, covering two workload types: \emph{document analysis} (60~tasks, 22~documents, ${\sim}$250~turns) and \emph{coding} (10~tasks). Document analysis is prefill-heavy (average ${\sim}$45K prompt tokens, only 984 decode tokens), while coding tasks generate longer outputs, making prefill a smaller fraction of wall time.

\autoref{tab:openclaw} reports the reduction in prompt tokens, wall time, and prefill latency.
On document analysis, \textsc{ContextPilot} reduces prompt tokens by \textbf{24.4\%} on average and \textbf{52.7\%} at P99, with prefill latency dropping by \textbf{63.6\%} and wall time by \textbf{20.7\%}.
On coding tasks, prefill savings remain large ($-62.2\%$), but wall-time reduction is modest ($-12.4\%$) since decode dominates.

\begin{table}[!t]
\centering
\caption{OpenClaw + SGLang with and without ContextPilot on \texttt{claw-tasks}.}
\label{tab:openclaw}
\resizebox{\linewidth}{!}{%
\begin{tabular}{llcccc}
\toprule
 & & \multicolumn{2}{c}{\textbf{Document Analysis}} & \multicolumn{2}{c}{\textbf{Coding}} \\
\cmidrule(lr){3-4} \cmidrule(lr){5-6}
\textbf{Metric} & \textbf{Method} & \textbf{Avg} & \textbf{P99} & \textbf{Avg} & \textbf{P99} \\
\midrule
\multirow{3}{*}{Prompt Tokens}
 & Baseline        & 45{,}771 & 92{,}785 & 32{,}118 & 38{,}952 \\
 & + ContextPilot  & 34{,}601 & 43{,}891 & 26{,}915 & 33{,}385 \\
 & $\Delta$        & $-24.4\%$ & $-52.7\%$ & $-16.2\%$ & $-14.3\%$ \\
\midrule
\multirow{3}{*}{Prefill Latency (s)}
 & Baseline        & 7.2 & 25.4 & 5.8 & 8.6 \\
 & + ContextPilot  & 2.6 & 6.7 & 2.2 & 4.9 \\
 & $\Delta$        & $-63.6\%$ & $-73.7\%$ & $-62.2\%$ & $-43.0\%$ \\
\midrule
\multirow{3}{*}{Wall Time (s)}
 & Baseline        & 26.1 & 68.8 & 19.5 & 28.7 \\
 & + ContextPilot  & 20.7 & 40.5 & 17.1 & 28.2 \\
 & $\Delta$        & $-20.7\%$ & $-41.2\%$ & $-12.4\%$ & $-1.7\%$ \\
\bottomrule
\end{tabular}}
\end{table}

\subsection{Deployment on Edge Devices}
To demonstrate that \textsc{ContextPilot}'s benefits extend beyond GPU servers, we evaluate Llama-3.2-1B-Instruct on two edge platforms using llama.cpp with batch size~1. \autoref{tab:edge} reports average latency on MultihopRAG. On the M3 MacBook Air (16\,GB), \textsc{ContextPilot} reduces latency from 3.31\,s to 1.38\,s ($2.41\times$); on the NVIDIA Jetson AGX Orin, latency drops from 2.12\,s to 1.41\,s ($1.50\times$). These results confirm that \textsc{ContextPilot}'s context reduction translates directly to wall-clock savings on resource-constrained edge devices.

\begin{table}[!t]
\centering
\caption{Llama-3.2-1B on edge devices with llama.cpp (MultihopRAG). ContextPilot achieves $1.5$--$2.4\times$ latency reduction across Apple Silicon laptops and the Jetson AGX Orin.}
\label{tab:edge}
\resizebox{\linewidth}{!}{%
\begin{tabular}{llc}
\toprule
\textbf{Device} & \textbf{Method} & \textbf{Avg Latency (s)} \\
\midrule
\multirow{2}{*}{M3 MacBook Air (16\,GB)}
 & llama.cpp                  & 3.31 \\
 & llama.cpp + ContextPilot   & \textbf{1.38} \\
\midrule
\multirow{2}{*}{NVIDIA Jetson AGX Orin}
 & llama.cpp                  &  2.12 \\
 & llama.cpp + ContextPilot   & \textbf{1.41} \\
\bottomrule
\end{tabular}}
\end{table}

\subsection{Performance breakdown, overhead and robustness}\label{subsec:eval-breakdown}

\begin{figure}[!t]
    \centering
    \begin{subfigure}[t]{0.49\linewidth}
        \centering
        \includegraphics[width=\linewidth]{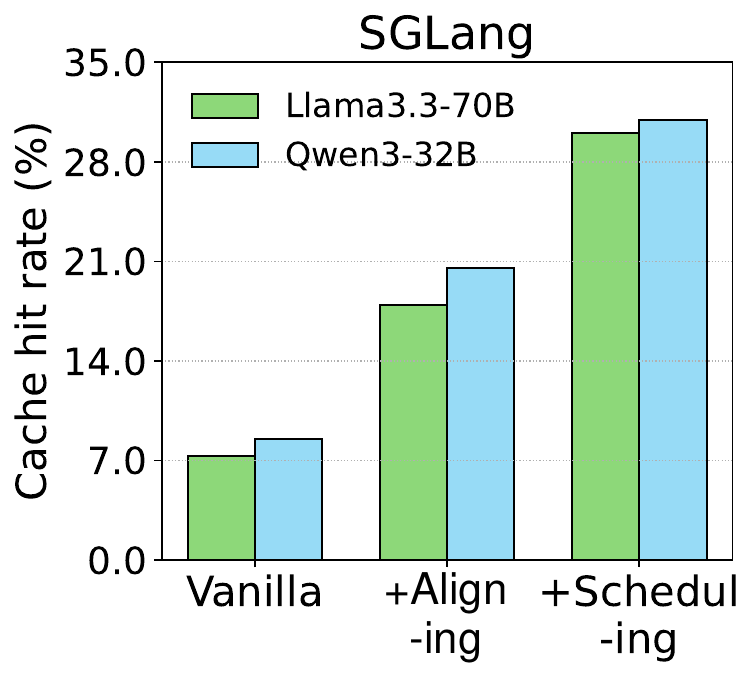}
        \label{fig:performance-llama}
    \end{subfigure}\hfill
    \begin{subfigure}[t]{0.49\linewidth}
        \centering
        \includegraphics[width=\linewidth]{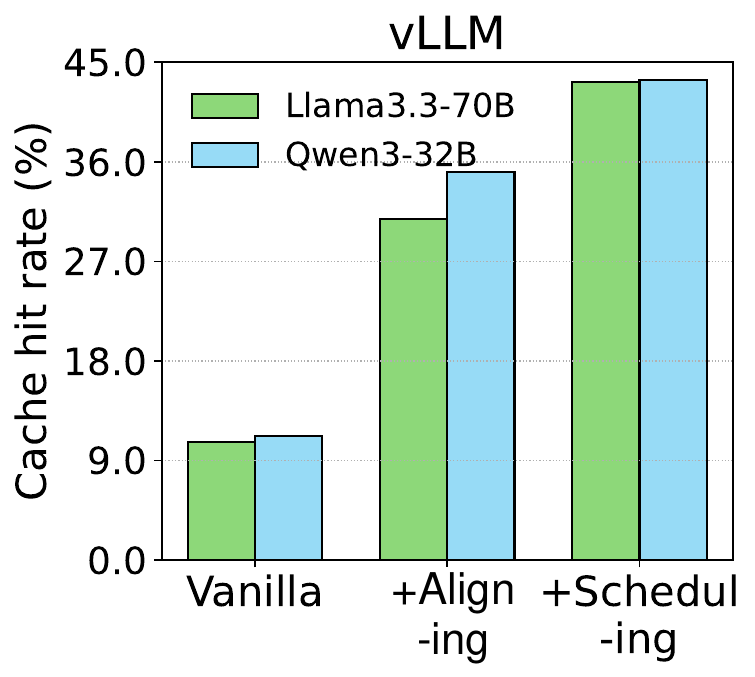}
        \label{fig:performance-qwen}
    \end{subfigure}
    \vspace{-0.5cm}
    \caption{Performance breakdown of key components.}
    \label{fig:break-h100}
    \vspace{-0.5cm}
\end{figure}

\mypar{Contribution of alignment and scheduling} We analyze how \emph{aligning} and \emph{scheduling} each contribute on MultihopRAG ($k{=}15$) across SGLang and vLLM on H100 GPUs. As shown in \autoref{fig:break-h100}, each component adds incremental cache hit ratio gains. For SGLang with Qwen3-32B, hit ratio rises from 8.49\% to 20.56\% (+ aligning) to 33.97\% (+ scheduling)—a 4× improvement. vLLM with Llama3.3-70B follows a similar trend: 10.7\% $\rightarrow$ 30.8\% $\rightarrow$ 43.2\%, directly translating to reduced prefill computation.

\mypar{Performance with long-running workloads}
Time-series analysis (\appref{apd:timeseries}) confirms that \textsc{ContextPilot}'s gains are sustained: it maintains ${\sim}$34\% cache hit ratio versus the baseline's ${\sim}$7\% (a consistent 5$\times$ advantage) throughout the entire workload, with both alignment and scheduling contributing to the improvement.

\mypar{Model accuracy under context reuse} A per-component breakdown (\appref{apd:accuracy-breakdown}) shows that alignment alone causes ${\leq}$1\% F1 variation, while adding annotations recovers this gap and even improves accuracy by +1.4--4.4\%.

\mypar{Overhead of context index construction} As shown in \autoref{tab:gpu-latency-multik}, index construction scales smoothly from 128 to 100K contexts on A6000 GPUs with minimal sensitivity to retrieval depth $k$. At 12K contexts, construction completes in 7.48\,s—negligible relative to total prefill latency. Even at 100K, \textsc{ContextPilot} finishes within 12~minutes, far faster than multi-hour offline prefilling. Per-request overhead is just ${\sim}$0.7~ms (\appref{apd:overhead}), negligible compared to seconds of prefill latency.

\begin{figure}[!t]
    \centering
        \includegraphics[width=\linewidth]{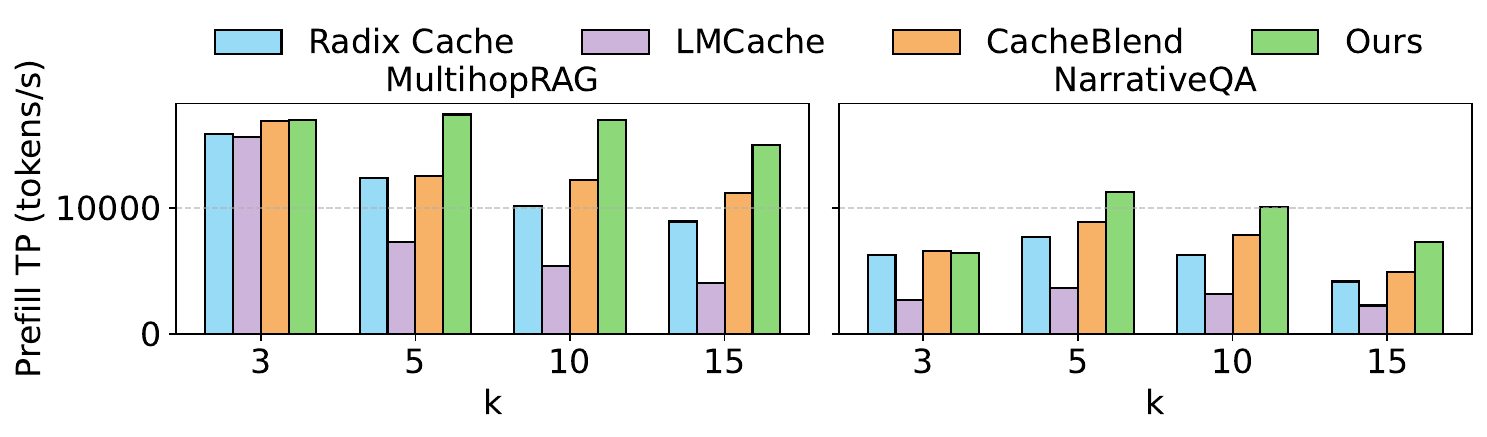}
        \vspace{-20pt}
        \caption{Prefill throughput under different top k values.}
        \label{fig:diff-k}
        \vspace{-0.4cm}
\end{figure}

\mypar{Performance with growing context lengths} We evaluate \textsc{ContextPilot} under varying retrieval depths (k=3,5,10,15) on NarrativeQA and MultihopRAG using A6000 GPUs. As shown in \autoref{fig:diff-k}, \textsc{ContextPilot} consistently achieves the highest prefill throughput across all k values. On MultihopRAG, it sustains 1.5–2.0× speedups over baselines as k grows from 3 to 15; on NarrativeQA, it maintains 1.3–1.6× gains.

\mypar{Effectiveness of context annotations} We tested placing annotations before and after questions, observing negligible impact on accuracy (variations below 0.5\%). Attention heatmap analysis further confirms that annotations effectively guide the model toward relevant documents, improving downstream accuracy. We include this analysis in the supplementary materials (\appref{apd:attn}).

\mypar{Impact of prefix cache size} Larger prefix caches benefit \textsc{ContextPilot} disproportionately: on MultihopRAG, SGLang hit ratios improve from 29.64\% to 33.97\% and vLLM from 35.90\% to 43.4\% when moving from A6000 (48\,GB) to H100 (80\,GB), while baselines see smaller gains. Details are in \appref{apd:cache-size}.

\newcommand{\twocolhead}[1]{\multicolumn{2}{c}{#1}}

\section{Related Works}
\mypar{RAG system optimization}  
System-level approaches such as METIS~\cite{metis} and Chameleon~\cite{chameleon} optimize workflow and hardware efficiency, jointly tuning retrieval settings or using heterogeneous accelerators~\cite{waferllm} to reduce latency and boost throughput. \textsc{ContextPilot} complements them by improving KV-cache reuse.

\mypar{Reranking in retrieval systems}  
Rerankers refine retrieval results via learned ranking models~\cite{adeyemi-etal-2024-zero,li2023making,qwen3embedding}; HyperRAG~\cite{hyperrag} further enables KV reuse at the reranking stage.  
\textsc{ContextPilot} instead operates downstream, aligning reranked document IDs with the prefix cache while preserving relevance through order annotations.

\mypar{Fine-tuning with positional re-encoding}  
Methods like BlockAttention, KVLink, and TurboRAG~\cite{blockattention,kvlink,turborag} fine-tune models to reuse KV caches via position re-encoding and pre-stored states.  
They improve efficiency but require heavy training and large cache storage.  
\textsc{ContextPilot} is training-free and shows strong promise in numerous RAG applications.

\mypar{Faster KV-cache compute}  
CacheBlend~\cite{cacheblend} and related works~\cite{cachegen,cachecraft,ape,alayadb,droidspeak,bitdecoding,marconi,tang2024quest,wu2025tokenselect,chen2025rapid,zhang2025pqcache} optimize KV-cache computation through compression, decoding, parallel encoding, cache sharing, or FLOP-aware admission and eviction policies.
\textsc{ContextPilot} complements these by operating at the context level, aligning and de-duplicating inputs to maximize reuse, and can be combined with them for further gains.

\section{Conclusion}
We presented \textsc{ContextPilot}, a context-reuse system that accelerates long-context prefill with negligible accuracy loss. \textsc{ContextPilot} uniformly represents external inputs as \emph{context blocks} and applies a common set of mechanisms---context indexing, alignment, de-duplication, and succinct annotations to boost prefix KV cache reuse across diverse workloads. Its modular design enables new system and algorithmic research on context engineering, management, and optimization for long-context AI. Looking ahead, we envision \textsc{ContextPilot} as a co-optimization framework that jointly decides \emph{what} context to feed and \emph{how} to serve it, maximizing both inference efficiency and output quality.

\section*{Acknowledgements}
We sincerely thank our shepherd and the MLSys reviewers for insightful feedback that significantly improved this manuscript. We also thank Minjie Wang (formerly Amazon AI Labs, now The University of Hong Kong (Shanghai X-Lab)) for valuable early-stage discussions, and Ryan Tsui (The University of Edinburgh) for contributions to the implementation. We further thank the Tencent WeChat (Weixin) Group for access to H20 GPU clusters for large-scale MoE evaluation and for insightful feedback. We also thank the UK Advanced Research and Invention Agency (ARIA) for funding this project. Finally, we thank the School of Informatics, University of Edinburgh, UK Isambard-AI and the Edinburgh International Data Facility (EIDF) for providing GPU resources that supported this research.

% \textbf{Do not} include acknowledgements in the initial version of
% the paper submitted for blind review.

% If a paper is accepted, the final camera-ready version can (and
% probably should) include acknowledgements. In this case, please
% place such acknowledgements in an unnumbered section at the
% end of the paper. Typically, this will include thanks to reviewers
% who gave useful comments, to colleagues who contributed to the ideas,
% and to funding agencies and corporate sponsors that provided financial

\nocite{langley00}

\bibliography{example_paper}
\bibliographystyle{mlsys2025}

%%%%%%%%%%%%%%%%%%%%%%%%%%%%%%%%%%%%%%%%%%%%%%%%%%%%%%%%%%%%%%%%%%%%%%%%%%%%%%%
%%%%%%%%%%%%%%%%%%%%%%%%%%%%%%%%%%%%%%%%%%%%%%%%%%%%%%%%%%%%%%%%%%%%%%%%%%%%%%%
% SUPPLEMENTAL CONTENT AS APPENDIX AFTER REFERENCES
%%%%%%%%%%%%%%%%%%%%%%%%%%%%%%%%%%%%%%%%%%%%%%%%%%%%%%%%%%%%%%%%%%%%%%%%%%%%%%%
%%%%%%%%%%%%%%%%%%%%%%%%%%%%%%%%%%%%%%%%%%%%%%%%%%%%%%%%%%%%%%%%%%%%%%%%%%%%%%%
% \newpage
\clearpage

\appendix
%%%%%%%%%%%%%%%%%%%%%%%%%%%%%%%%%%%%%%%%%%%%%%%%%%%%%%%%%%%%%%%%%%%%%%%%%%%%%%%
% SUPPLEMENTAL CONTENT AS APPENDIX AFTER REFERENCES
%%%%%%%%%%%%%%%%%%%%%%%%%%%%%%%%%%%%%%%%%%%%%%%%%%%%%%%%%%%%%%%%%%%%%%%%%%%%%%%

\section{DeepSeek-R1 Results}
\label{apd:deepseek-r1}

\autoref{tab:deepseek-r1} presents end-to-end results for DeepSeek-R1 on MultihopRAG and NarrativeQA datasets, evaluated on $16\times$H20 and $32\times$H20 GPUs. ContextPilot achieves $1.81\times$ and $1.52\times$ throughput improvements on MultihopRAG and NarrativeQA respectively, with these speedups remaining consistent across both 16 and 32 GPU deployments using context-aware routing.

\begin{table}[H]
\centering
\caption{DeepSeek-R1 results formatted for vertical readability. By stacking hardware configurations, the table width is minimized while preserving all data points.}
\label{tab:deepseek-r1}
\resizebox{\linewidth}{!}{%
\begin{tabular}{llccc}
\toprule
\textbf{Method} & \textbf{Hardware} & \textbf{Prefill TP (tok/s)} & \textbf{Cache Hit} & \textbf{F1 (\%)} \\
\midrule
\multicolumn{5}{c}{\textit{\textbf{Dataset: MultihopRAG}}} \\
\midrule
\multirow{2}{*}{Vanilla}
 & 16$\times$H20 & 9636.69 & 5.12\% & 64.15 \\
 & 32$\times$H20 & 18406.08 & 4.17\% & 64.15 \\
\cmidrule{1-5}
\multirow{2}{*}{ContextPilot w/o Annotations}
 & 16$\times$H20 & 17498.75 & 60.37\% & 64.09 \\
 & 32$\times$H20 & 33072.64 & 58.41\% & 64.09 \\
\cmidrule{1-5}
\multirow{2}{*}{ContextPilot (Ours)}
 & 16$\times$H20 & \textbf{17498.75} & \textbf{60.37\%} & \textbf{64.68} \\
 & 32$\times$H20 & \textbf{33072.64} & \textbf{58.41\%} & \textbf{64.68} \\

\midrule
\multicolumn{5}{c}{\textit{\textbf{Dataset: NarrativeQA}}} \\
\midrule
\multirow{2}{*}{Vanilla}
 & 16$\times$H20 & 8687.98 & 6.08\% & 40.20 \\
 & 32$\times$H20 & 16247.32 & 5.14\% & 40.20 \\
\cmidrule{1-5}
\multirow{2}{*}{ContextPilot w/o Annotations}
 & 16$\times$H20 & 13201.52 & 38.24\% & 40.38 \\
 & 32$\times$H20 & 24686.84 & 35.71\% & 40.38 \\
\cmidrule{1-5}
\multirow{2}{*}{ContextPilot (Ours)}
 & 16$\times$H20 & \textbf{13201.52} & \textbf{38.24\%} & \textbf{41.08} \\
 & 32$\times$H20 & \textbf{24686.84} & \textbf{35.71\%} & \textbf{41.08} \\
\bottomrule
\end{tabular}}
\end{table}

%%%%%%%%%%%%%%%%%%%%%%%%%%%%%%%%%%%%%%%%%%%%%%%%%%%%%%%%%%%%%%%%%%%%%%%%%%%%%%%
\section{Attention Map Analysis}
\label{apd:attn}
Since context engineering~\cite{anthropic2025contextengineering} and in-context learning~\cite{kirsch2022general} strongly influence model inference, we analyze attention patterns when explicit annotations are introduced to recall the original relevance ranking.

\begin{figure}[H]
    \centering
    \includegraphics[width=\linewidth]{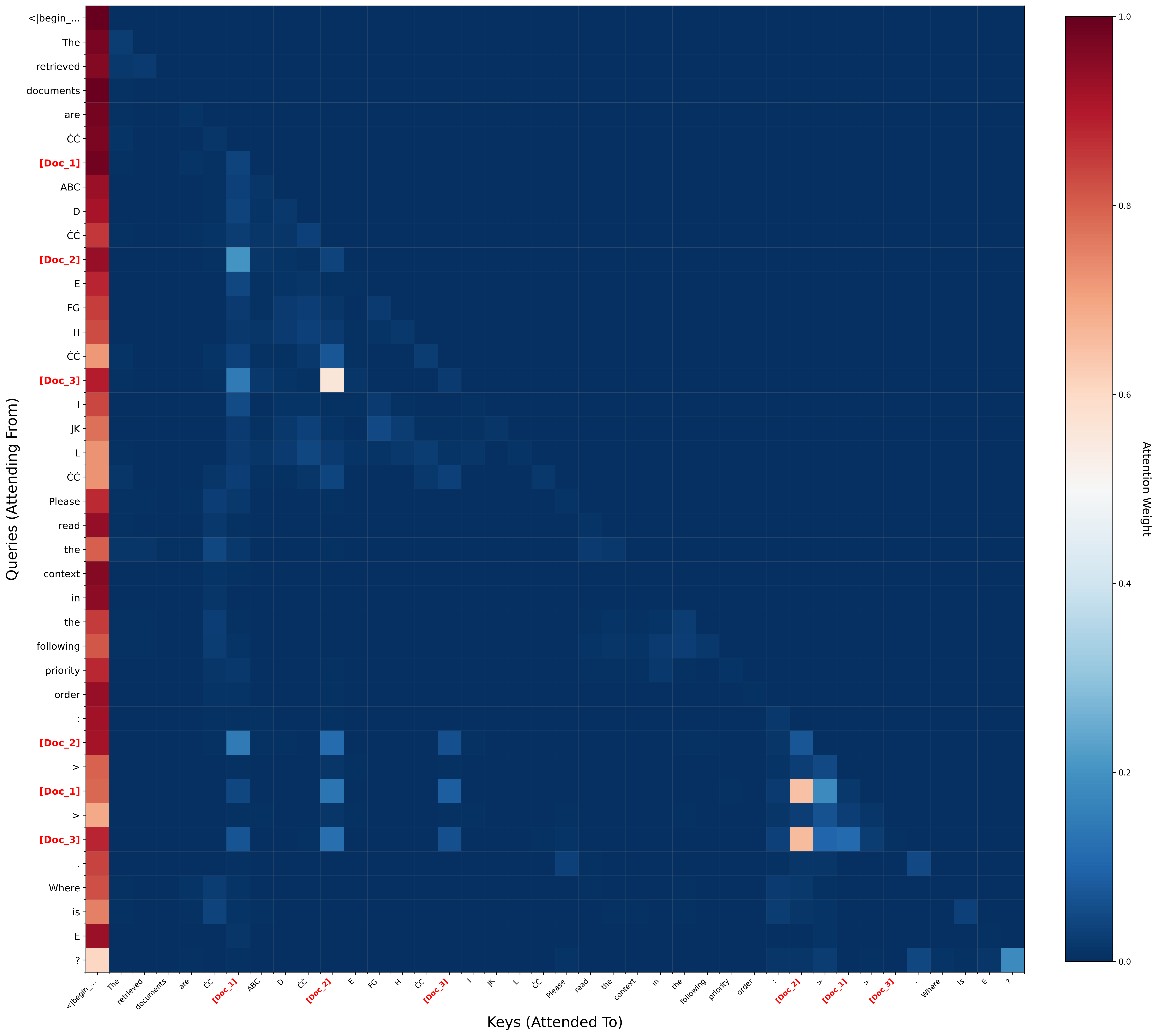}
    \caption{Attention map of the last layer attention of LLaMA3.3 (Head 14) for the prompt ``The retrieved documents are [Doc\_1] ABCD  [Doc\_2] EFGH  [Doc\_3] IJKL. Please read the context in the following priority order: [Doc\_2] $>$ [Doc\_1] $>$ [Doc\_3]. Where is E?''.}
    \label{fig:attn-llama}
\end{figure}

\begin{figure}[H]
    \centering
    \includegraphics[width=\linewidth]{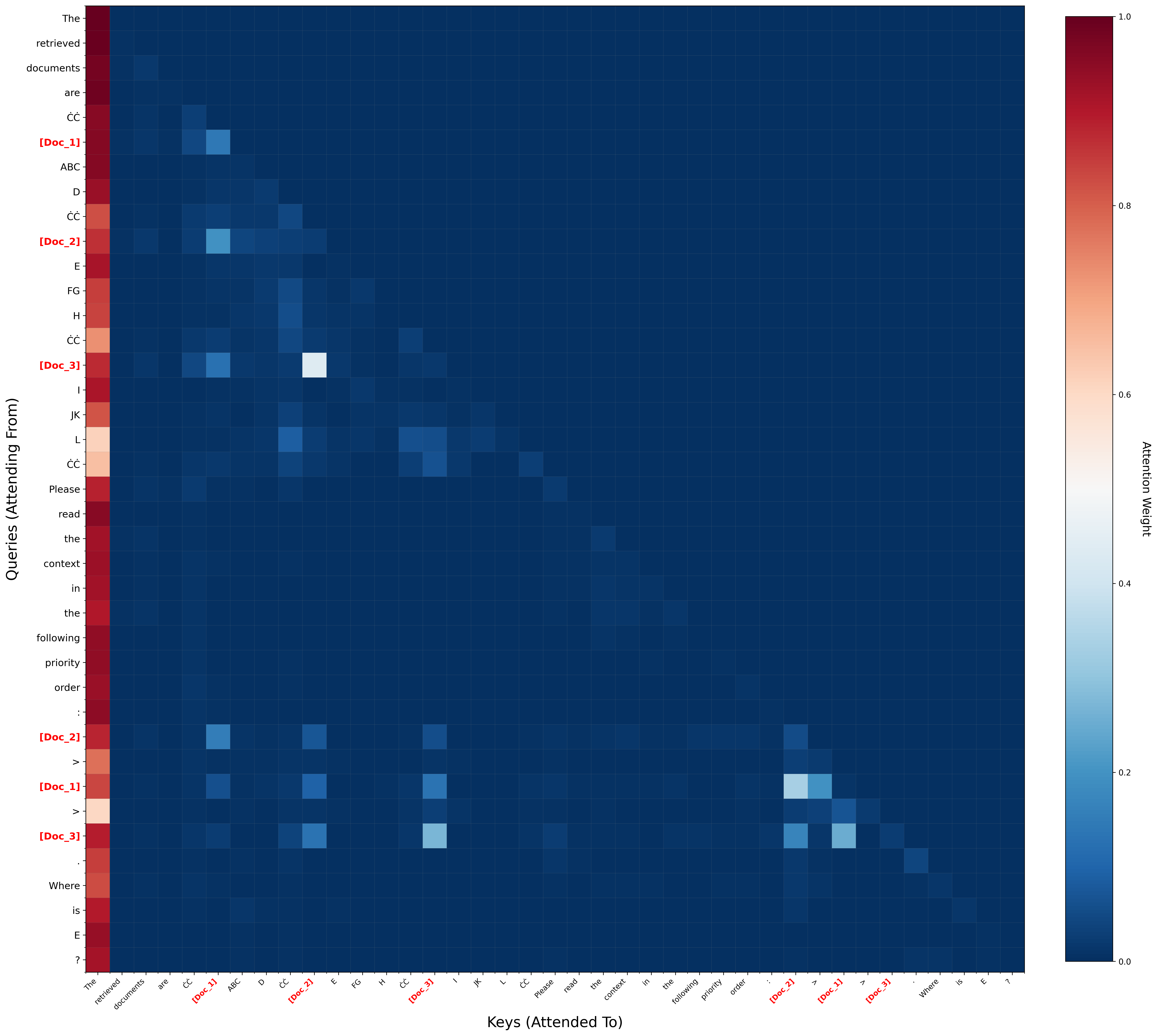}
    \caption{Attention map of the last layer attention of Qwen3 (Head 9) for the prompt ``The retrieved documents are [Doc\_1] ABCD  [Doc\_2] EFGH  [Doc\_3] IJKL. Please read the context in the following priority order: [Doc\_2] $>$ [Doc\_1] $>$ [Doc\_3]. Where is E?''.}
    \label{fig:attn-qwen}
\end{figure}

\autoref{fig:attn-llama} and~\autoref{fig:attn-qwen} compare the final-layer attention maps of Qwen3 and LLaMA3.3 under this setup. When given explicit document-priority cues, both models exhibit consistent attention behaviors despite architectural differences. They correctly focus on document tokens ([Doc\_1], [Doc\_2], [Doc\_3]), reflecting awareness of the mismatch between the aligned and original sequences, as indicated by intersections between queries in the annotation region and keys in the context region. As the context re-aligns with the original sequence, both models emphasize ([Doc\_2]) while parsing ([Doc\_1]) and ([Doc\_3]), showing that the cue ([Doc\_2] $>$ [Doc\_1] $>$ [Doc\_3]) effectively directs cross-document attention. Hence, explicit annotations reshape internal attention, aligning it with semantic rather than positional priority.

This finding supports a central hypothesis of ContextPilot: explicit annotations can make a comeback from the accuracy lost to alignment and filtering by re-establishing alignment with the original retrieval semantics.

%%%%%%%%%%%%%%%%%%%%%%%%%%%%%%%%%%%%%%%%%%%%%%%%%%%%%%%%%%%%%%%%%%%%%%%%%%%%%%%
\section{Document Access Distribution}
\label{apd:cdf}

\begin{figure}[H]
    \centering
    \includegraphics[width=0.9\linewidth]{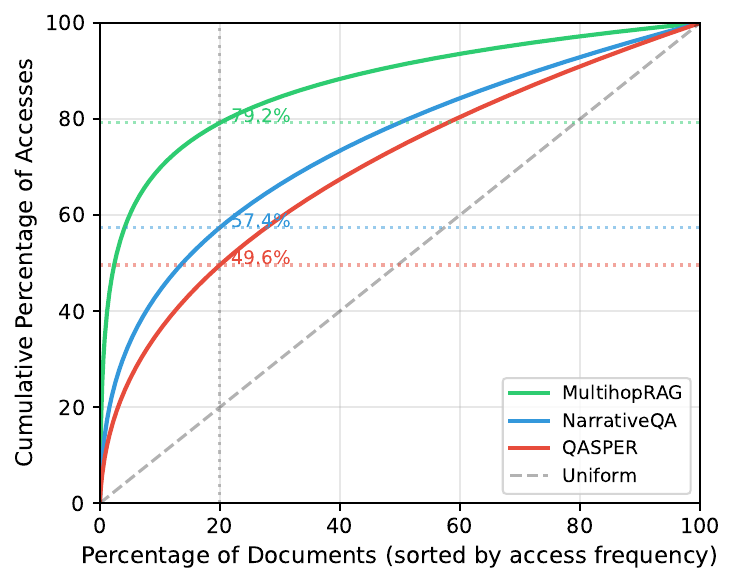}
    \caption{Document access distribution (CDF) across three datasets. The vertical dashed line marks 20\% of documents; horizontal lines indicate the cumulative percentage of accesses from these top documents.}
    \label{fig:cdf}
\end{figure}

\autoref{fig:cdf} shows the cumulative distribution of document access frequency across three RAG datasets. A small fraction of documents accounts for the majority of retrievals: the top 20\% most frequently accessed documents cover 79.2\% of retrieval events on MultihopRAG, 57.4\% on NarrativeQA, and 49.6\% on QASPER. This heavy-tailed distribution confirms that real-world RAG workloads exhibit substantial context overlap across sessions, motivating context reuse through prefix-aligned alignment.

%%%%%%%%%%%%%%%%%%%%%%%%%%%%%%%%%%%%%%%%%%%%%%%%%%%%%%%%%%%%%%%%%%%%%%%%%%%%%%%
\section{Additional Evaluation Details}
\label{apd:additional-eval}

%
% \end{figure}

\subsection{Time-Series Metrics}
\label{apd:timeseries}

\begin{figure}[H]
    \centering
    \includegraphics[width=\linewidth]{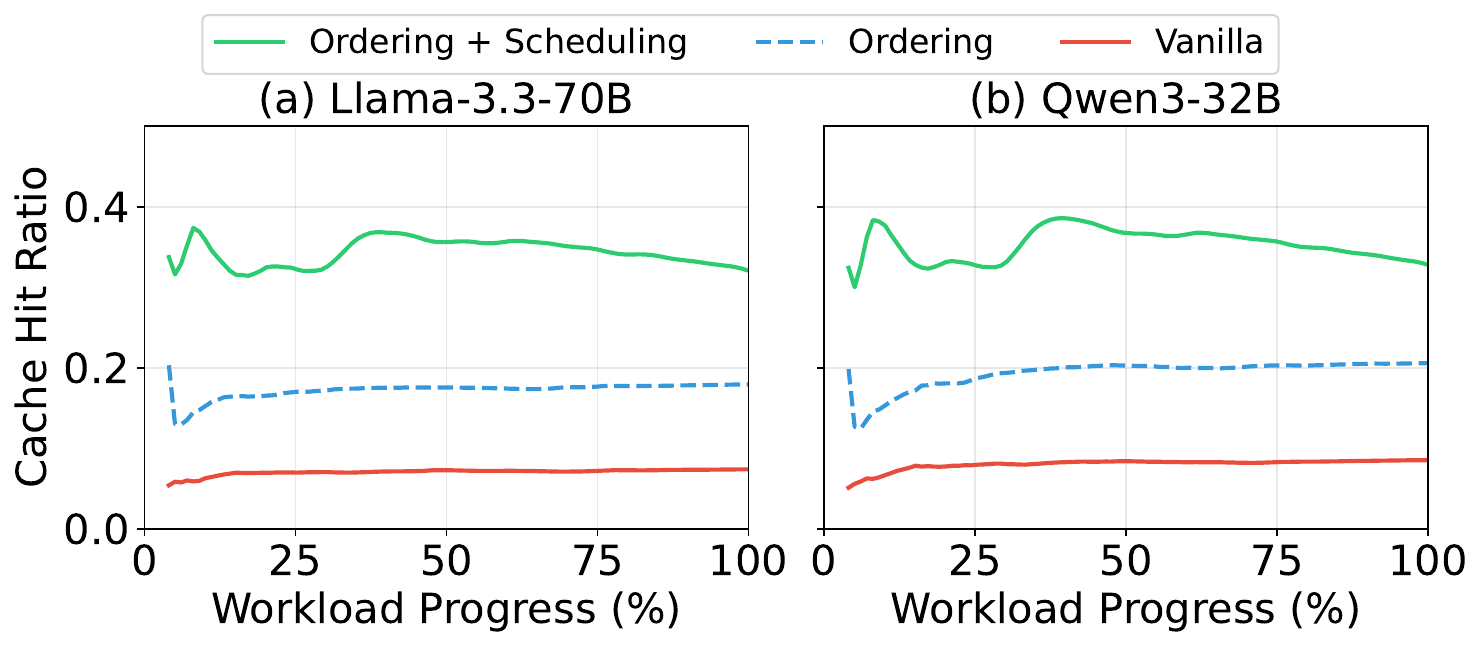}
    \caption{Cache hit ratio over workload progress for Llama-3.3-70B and Qwen3-32B. ContextPilot maintains $\sim$5$\times$ higher hit ratio throughout execution.}
    \label{fig:hit-ratio-time}
\end{figure}

\autoref{fig:hit-ratio-time} shows how cache hit ratio evolves as the workload progresses. ContextPilot maintains approximately 34\% cache hit ratio compared to baseline's 7\% throughout the entire workload, demonstrating a sustained 5$\times$ improvement that is not a transient warm-up effect.

\begin{figure}[H]
    \centering
    \includegraphics[width=\linewidth]{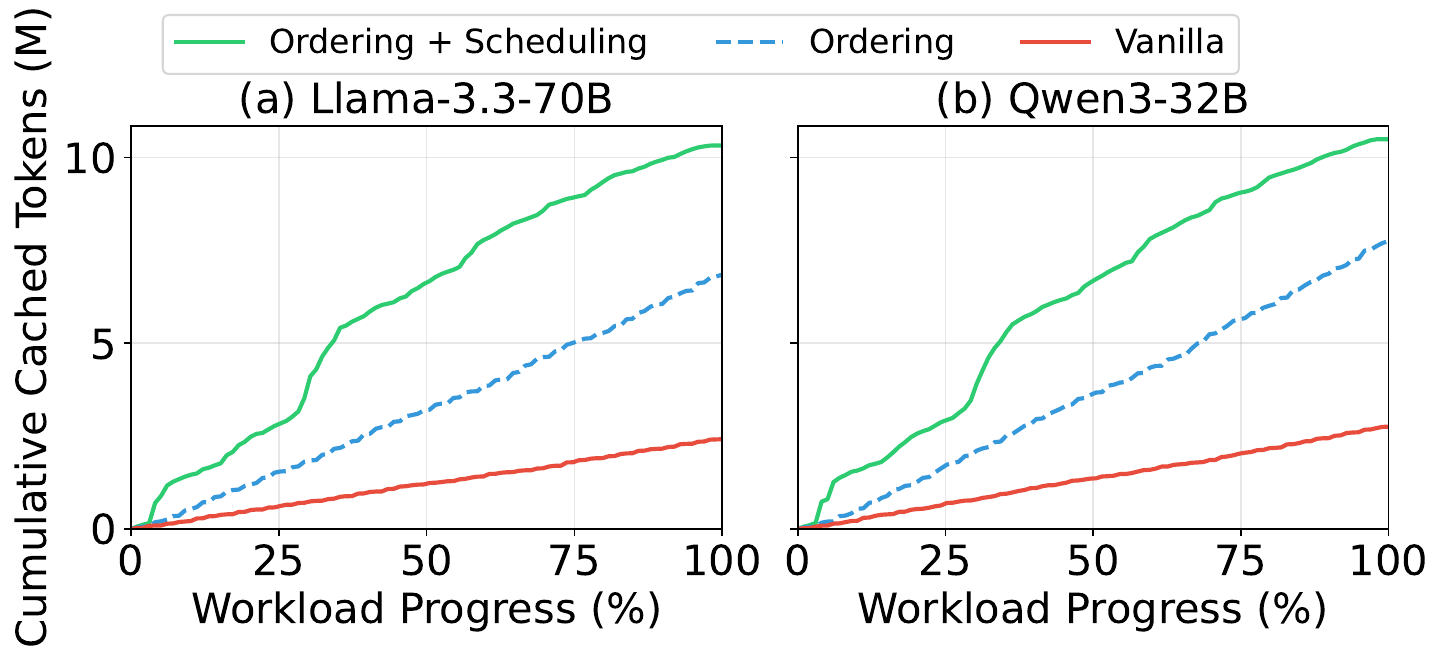}
    \caption{Cumulative cached tokens (radix tree reuse) over workload progress for Llama-3.3-70B and Qwen3-32B. ContextPilot achieves 4$\times$ better prefix reuse.}
    \label{fig:radix-tree-util}
\end{figure}

\autoref{fig:radix-tree-util} presents cumulative cached tokens as a metric for radix tree prefix reuse. ContextPilot achieves 10.33M cached tokens versus baseline's 2.42M at completion on Llama3.3-70B-Instruct (4.27$\times$), and 10.50M versus 2.75M on Qwen3-32B (3.82$\times$). The ``w/o Scheduling'' variant (6.85M, 2.83$\times$) confirms that both alignment and scheduling contribute to the improvement.

\subsection{Accuracy Breakdown}
\label{apd:accuracy-breakdown}

\autoref{tab:accuracy-breakdown} provides a detailed breakdown of accuracy contributions from each ContextPilot component.

\begin{table}[H]
\centering
\caption{Accuracy breakdown by component. Alignment alone: ${\leq}$1\% F1 variation; adding annotations: +1.4--4.4\% gains over aligned baseline.}
\label{tab:accuracy-breakdown}
\resizebox{\linewidth}{!}{%
\begin{tabular}{ll|cc}
\toprule
\textbf{Model} & \textbf{Configuration} & \textbf{MultihopRAG} & \textbf{NarrativeQA} \\
\midrule
\multirow{4}{*}{Qwen3-32B}
& Baseline      & 60.4\% & 28.4\% \\
& + Alignment  & 60.0\% & 28.2\% \\
& + Annotation        & 64.4\% & 29.6\% \\
& + Scheduling  & \textbf{64.4\%} & \textbf{29.6\%} \\
\midrule
\multirow{4}{*}{Qwen3-4B}
& Baseline      & 35.2\% & 16.0\% \\
& + Alignment  & 34.5\% & 15.2\% \\
& + Annotation        & 36.6\% & 17.3\% \\
& + Scheduling  & \textbf{36.6\%} & \textbf{17.3\%} \\
\bottomrule
\end{tabular}}
\end{table}

\subsection{Per-Request Overhead}
\label{apd:overhead}

\autoref{tab:overhead} reports the per-request overhead of ContextPilot components, measured on 2K requests with $k=15$ on NVIDIA A6000.

\begin{table}[H]
\centering
\caption{Per-request overhead breakdown. Total overhead ($\sim$0.7 ms) is negligible compared to seconds of prefill latency.}
\label{tab:overhead}
\begin{tabular}{lc}
\toprule
\textbf{Component} & \textbf{Latency (ms)} \\
\midrule
Search          & 0.068 \\
Alignment       & 0.047 \\
De-duplication  & 0.600 \\
\midrule
\textbf{Total}  & \textbf{$\sim$0.7} \\
\bottomrule
\end{tabular}
\end{table}

%%%%%%%%%%%%%%%%%%%%%%%%%%%%%%%%%%%%%%%%%%%%%%%%%%%%%%%%%%%%%%%%%%%%%%%%%%%%%%%
\section{OpenClaw Pipeline Diagram}
\label{apd:openclaw}

\begin{figure}[H]
    \centering
    \includegraphics[width=\linewidth]{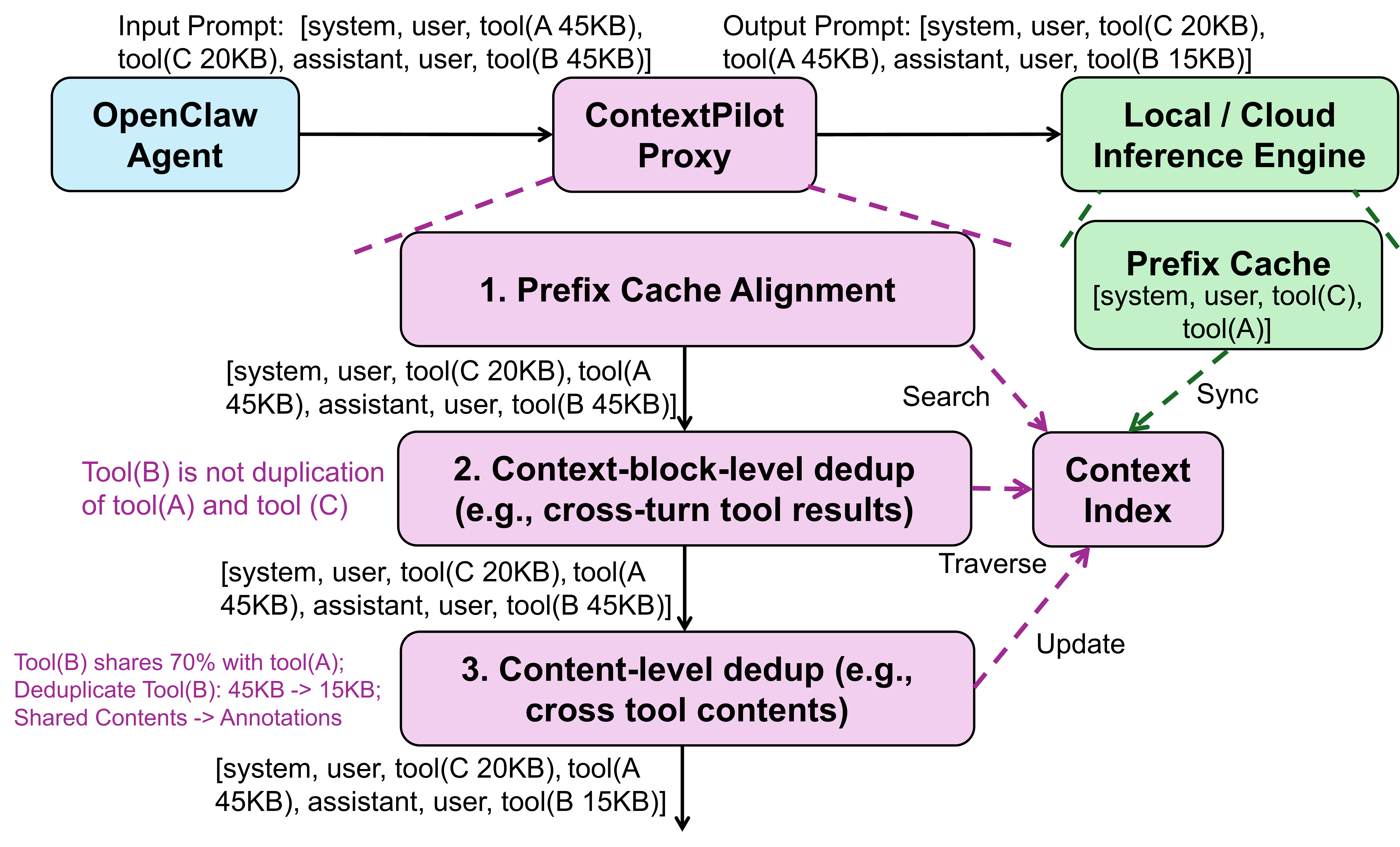}
    \caption{The OpenClaw + ContextPilot pipeline. The OpenClaw agent sends requests to the ContextPilot proxy, which aligns prompts with the prefix cache of the inference engine, deduplicates overlapping context both across turns and within context blocks, and forwards optimized prompts to the inference engine.}
    \label{fig:openclaw}
\end{figure}

%%%%%%%%%%%%%%%%%%%%%%%%%%%%%%%%%%%%%%%%%%%%%%%%%%%%%%%%%%%%%%%%%%%%%%%%%%%%%%%
\section{System Overhead with Zero Context Overlap}
\label{apd:zero-overlap}

Zero context overlap represents the worst case for ContextPilot, as it isolates pure system overhead with no reuse benefit. Using a synthetic RAG workload with no retrieval overlap, ContextPilot adds only 0.72~s of prefill latency for 1K contexts (one-hour job), demonstrating that querying the context index during operation incurs negligible overhead.

%%%%%%%%%%%%%%%%%%%%%%%%%%%%%%%%%%%%%%%%%%%%%%%%%%%%%%%%%%%%%%%%%%%%%%%%%%%%%%%
\section{Impact of Prefix Cache Size}
\label{apd:cache-size}

The prefix KV-cache stores precomputed attention states for reuse across requests; a larger cache retains more contexts, increasing the chance of a cache hit and reducing redundant prefill computation. We evaluate this effect on MultihopRAG across A6000 (48\,GB) and H100 (80\,GB) GPUs. Because ContextPilot aligns contexts with the prefix cache to maximize overlap, it benefits disproportionately from larger caches: SGLang hit ratios improve from 29.64\% to 33.97\%, and vLLM from 35.90\% to 43.4\%, translating directly into higher prefill throughput. In contrast, baselines see smaller gains since their context alignment does not systematically exploit the additional capacity.

\section{Detailed Algorithm Pseudocode}
\label{apd:algorithms}

This appendix collects all algorithm pseudocode referenced in the main text. \autoref{alg:search} details the context index tree search, \autoref{alg:context-ordering} describes context alignment for prefix sharing, \autoref{alg:dedup} formalizes context de-duplication, \autoref{alg:cluster-reorder} provides the full pseudocode for context index construction via hierarchical clustering, and \autoref{alg:grouping} details the tree-based request grouping and scheduling procedure.

\begin{algorithm}[!ht]
\caption{Context Index Tree Search}
\label{alg:search}
\begin{algorithmic}[1]
\REQUIRE Context $C = \{b_1, \ldots, b_k\}$, root node $R$
\ENSURE Search path, best matching node
\STATE $cur \gets R$, $path \gets []$
\WHILE{$cur$ has children}
    \STATE $best \gets \arg\min_{c \in cur.\text{children},\, C \cap c.\text{blocks} \neq \emptyset} \textsc{Dist}(C, c)$
    \IF{no overlapping child found}
        \STATE \textbf{break}
    \ENDIF
    \STATE $path.\textsc{append}(\text{index of } best)$
    \IF{$best$ is leaf}
        \STATE \textbf{return} $path$, $best$
    \ENDIF
    \STATE $cur \gets best$
\ENDWHILE
\STATE \textbf{return} $path$, $cur$
\end{algorithmic}
\end{algorithm}

\begin{algorithm}[!ht]
\caption{Context Alignment for Prefix Sharing}
\label{alg:context-ordering}
\begin{algorithmic}[1]
\REQUIRE Context $C$ with context blocks, Context Tree root node $R$
\ENSURE Aligned context $C'$
    \STATE $bestMatch \gets \textsc{FindBestMatchNode}(C, R)$
    \IF{$bestMatch = \text{null}$}
        \STATE $C' \gets C$
        \STATE \textbf{return} $C'$
    \ENDIF
    \STATE $prefix \gets bestMatch.context$
    \STATE $remaining \gets C \setminus prefix$
    \STATE $C' \gets prefix \cup remaining$
    \STATE \textbf{return} $C'$
\vspace{0.5em}
\STATE \textbf{function} \textsc{FindBestMatchNode}($C$, $R$)
    \IF{$C$ is initialization context}
        \STATE \textbf{return} $C.parent$
    \ELSE
        \STATE \textbf{return} $\textsc{SearchTree}(C, R)$
    \ENDIF
\STATE \textbf{end function}
\end{algorithmic}
\end{algorithm}

\begin{algorithm}[!ht]
\caption{Context De-duplication}
\label{alg:dedup}
\begin{algorithmic}[1]
\REQUIRE New context $C_{\text{new}}$, context index $\mathcal{I}$, conversation ID $id$, chunk modulus $M$
\ENSURE Deduplicated context $C'$, location annotations $H$
\STATE $S \gets \mathcal{I}.\text{seen\_blocks}[id]$ \hfill $\triangleright$ blocks indexed in prior turns
\STATE $V \gets \mathcal{I}.\text{seen\_subblocks}[id]$ \hfill $\triangleright$ sub-block hashes from prior turns
\STATE $C' \gets []$, $H \gets []$
\FOR{each block $b$ in $C_{\text{new}}$}
    \IF{$b \in S$}
        \STATE $h \gets \text{``refer to } [b] \text{ in previous conversation''}$
        \STATE $C'.\textsc{append}(h)$; $H.\textsc{append}(h)$
    \ELSE
        \STATE $\{s_1, \ldots, s_m\} \gets \textsc{CDC}(b, M)$ \hfill $\triangleright$ content-defined chunking
        \FOR{$j = 1$ to $m$}
            \STATE $f \gets \textsc{Hash}(s_j)$
            \IF{$f \in V$ \AND $V[f] \neq b$}
                \STATE Replace $s_j$ with location annotation to $V[f]$
            \ELSE
                \STATE $V[f] \gets b$
            \ENDIF
        \ENDFOR
        \STATE $C'.\textsc{append}(\textsc{Concat}(s_1, \ldots, s_m))$
    \ENDIF
\ENDFOR
\STATE $S \gets S \cup C_{\text{new}}$ \hfill $\triangleright$ register for future turns
\STATE $\mathcal{I}.\text{seen\_subblocks}[id] \gets V$ \hfill $\triangleright$ persist sub-block hashes
\STATE \textbf{return} $C'$, $H$
\end{algorithmic}
\end{algorithm}

\begin{algorithm}[!ht]
\caption{Context Index Construction via Hierarchical Clustering}
\label{alg:cluster-reorder}
\textbf{Input:} Batch of $n$ contexts $\mathcal{S} = \{s_1, s_2, \ldots, s_n\}$, distance function $d(\cdot,\cdot)$

\begin{algorithmic}[1]
\STATE \textbf{// Phase 1: Distance computation and clustering}
\STATE $D \gets$ pairwise distances using $d(s_i, s_j)$ \hfill $\triangleright$ GPU or CPU
\STATE $Z \gets \textsc{Linkage}(D)$ \hfill $\triangleright$ hierarchical clustering

\STATE \textbf{// Phase 2: Build tree with deduplication}
\FOR{$i = 1$ to $n$}
    \STATE Create leaf node $v_i$ with $v_i.\text{content} \gets s_i$
    \IF{$\exists\, v_j$ with $v_j.\text{content} = s_i$}
        \STATE Redirect $v_i \to v_j$; $v_j.\text{freq} \mathrel{+}= 1$ \hfill $\triangleright$ dedup
    \ENDIF
\ENDFOR
\FOR{each merge $(c_1, c_2, \delta)$ in $Z$}
    \STATE $v \gets$ new internal node
    \STATE $v.\text{content} \gets c_1.\text{content} \cap c_2.\text{content}$
    \STATE $v.\text{children} \gets [c_1, c_2]$; \; $c_1.\text{parent}, c_2.\text{parent} \gets v$
\ENDFOR
\STATE Remove empty internal nodes; relink children to grandparents

\STATE \textbf{// Phase 3: Top-down prefix alignment}
\STATE Compute search paths from root to all nodes
\FOR{each node $v$ in BFS order from root}
    \IF{$v$ is not root \textbf{and} $v.\text{parent}.\text{docs} \neq \emptyset$}
        \STATE $v.\text{docs} \gets v.\text{parent}.\text{docs} \oplus (v.\text{docs} \setminus v.\text{parent}.\text{docs})$
    \ENDIF
\ENDFOR
\STATE \textbf{return} aligned contexts from leaf nodes, search paths
\end{algorithmic}
\end{algorithm}

\begin{algorithm}[!ht]
\caption{Search-Path-Based Request Grouping and Scheduling}
\label{alg:grouping}
\begin{algorithmic}[1]
\STATE \textbf{Input:} $N$ contexts with search paths $\{P_1, \ldots, P_N\}$ from alignment
\STATE \textbf{Output:} Scheduled execution order
\\
\STATE \textbf{// Phase 1: Group by root prefix} \hfill $\triangleright$ $O(N)$
\STATE $\mathcal{G} \gets$ empty map
\FOR{$i = 1$ to $N$}
    \STATE $key \gets P_i[0]$ \hfill $\triangleright$ first element of search path
    \STATE $\mathcal{G}[key].\textsc{append}(i)$
\ENDFOR
\\
\STATE \textbf{// Phase 2: Sort within each group} \hfill $\triangleright$ $O(N \log N)$
\FOR{each group $G \in \mathcal{G}$}
    \STATE Sort $G$ by $|P_i|$ descending \hfill $\triangleright$ longer paths first
\ENDFOR
\\
\STATE \textbf{// Phase 3: Order groups and flatten}
\STATE Sort groups by $|G|$ descending \hfill $\triangleright$ largest groups first
\STATE \textbf{return} concatenation of all groups
\end{algorithmic}
\end{algorithm}

\end{sloppypar}
\end{document}